%% file: acl_latex.tex
\documentclass[11pt]{article}

\usepackage[table]{xcolor}

\usepackage[final]{acl}

\usepackage{times}
\usepackage{latexsym}

\usepackage[T1]{fontenc}

\usepackage[utf8]{inputenc}

\usepackage{microtype}

\usepackage{inconsolata}

\usepackage{graphicx}
\usepackage{multirow}
\usepackage{pifont}%
\usepackage{makecell}
\usepackage[skins]{tcolorbox}
\tcbuselibrary{breakable}
\usepackage[switch]{lineno}
\usepackage{booktabs}
\usepackage{amssymb}
\usepackage{enumitem}
\usepackage{amsmath}
\usepackage{url}
\usepackage[ruled,vlined]{algorithm2e}
\usepackage{mdframed}
\usepackage{quoting}
\quotingsetup{vskip=0pt, leftmargin=10pt, rightmargin=10pt}
\usepackage{subcaption}

\definecolor{lightblue}{RGB}{220,220,255}
\definecolor{checkmarkgreen}{RGB}{34,139,34}
\definecolor{crossmarkred}{RGB}{220,20,60}
\newcommand{\cmark}{\textcolor{checkmarkgreen}{\ding{51}}}
\newcommand{\xmark}{\textcolor{crossmarkred}{\ding{55}}}

\tcbset{
  mybox/.style={
    colframe=purple!65!black,          %
    colback=purple!5!white,          %
    coltitle=white,              %
    colbacktitle=purple!65!black, %
    fonttitle=\bfseries,         %
    boxrule=1pt,                 %
    toptitle=2mm,                %
    bottomtitle=2mm,             %
    arc=4mm,                  %
    before=\vskip10pt,           %
    after=\vskip10pt             %
  }
}

\title{Structured Preference Optimization for Vision-Language Long-Horizon Task Planning}

\author{Xiwen Liang$^1$$^*$, Min Lin$^1$$^*$, Weiqi Ruan$^2$, Rongtao Xu$^{3}$, Yuecheng Liu$^4$, \\ \textbf{Jiaqi Chen$^5$, Bingqian Lin$^6$, Yuzheng Zhuang$^4$, Xiaodan Liang$^1$$^\dagger$} \\
$^1$Shenzhen Campus of Sun Yat-sen University, $^2$The Chinese University of Hong Kong, \\
$^3$Spatialtemporal AI, $^4$Huawei Noah's Ark Lab, $^5$The University of Hong Kong, \\
$^6$Shanghai Jiaotong University
}

\begin{document}
\maketitle

\let\thefootnote\relax\footnotetext{$^*$Equal contribution.}
\footnotetext{$^\dagger$Corresponding author.}

\input{sec/0_abstract}    
\input{sec/1_intro}
\input{sec/2_related_work}
\input{sec/3_method}
\input{sec/4_benchmark}

\input{sec/5_experiments}

\input{sec/6_conclusion}

\bibliography{custom}

\appendix

\input{sec/X_suppl}

\end{document}

%% file: sec/0_abstract.tex
\begin{abstract}
Existing vision-language planning methods perform well on short-horizon tasks but struggle with long-horizon reasoning in dynamic environments due to the difficulty of training models to generate high-quality reasoning processes. To address this, we propose \textbf{S}tructured \textbf{P}reference \textbf{O}ptimization (\textbf{SPO}), a framework that enhances reasoning and action selection for long-horizon task planning through structured evaluation and optimized training.
SPO introduces:
1) Structured Preference Evaluation and Optimization, which evaluates reasoning chains across task relevance, historical consistency (as part of textual coherence), and image awareness (alignment with visual observations) to construct high-quality preference pairs;
and 2) Curriculum-Guided Progressive Learning, enabling the model to adapt from simple to complex tasks, thereby improving generalization and robustness.
To advance research in vision-language long-horizon task planning, we introduce ExtendaBench, a comprehensive benchmark covering 1,509 tasks across VirtualHome and Habitat 2.0, categorized into ultra-short, short, medium, and long tasks.
Experimental results demonstrate that SPO significantly improves reasoning quality and final decision accuracy, outperforming prior methods on long-horizon tasks and underscoring the effectiveness of preference-driven optimization in vision-language task planning.
Specifically, SPO achieves a +5.98\% GCR and +4.68\% SR improvement in VirtualHome and a +3.30\% GCR and +2.11\% SR improvement in Habitat over the best-performing baselines.
\end{abstract}

%% file: sec/1_intro.tex
\section{Introduction}

\begin{figure}
    \centering
    \includegraphics[width=\linewidth]{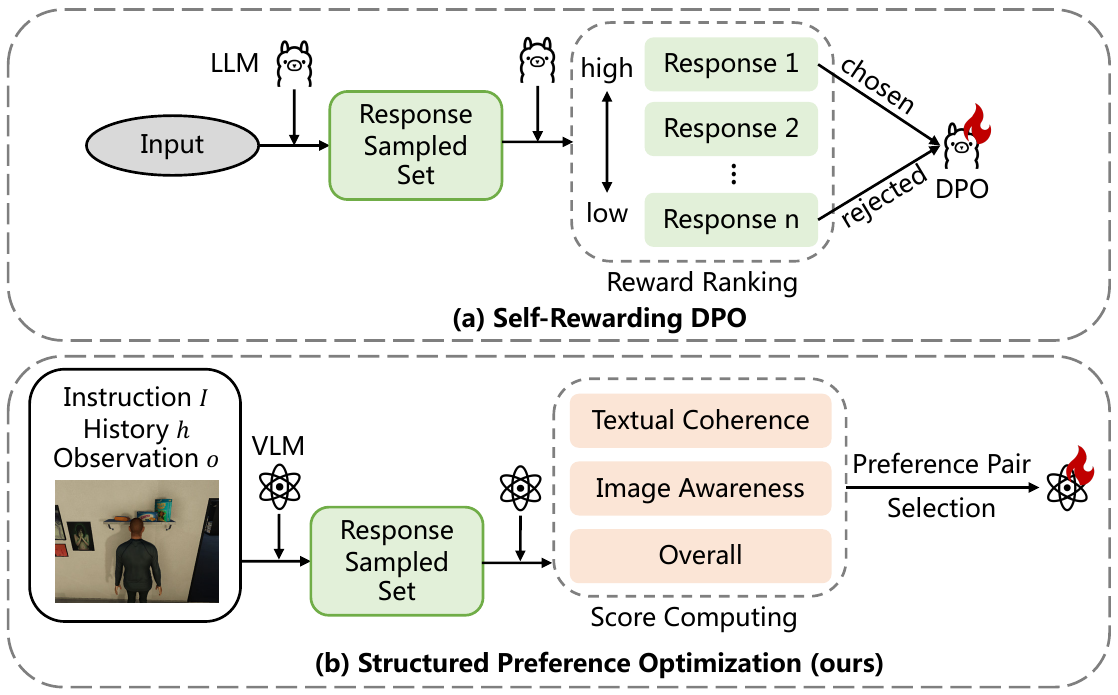}
    \vspace{-4mm}
    \caption{Comparison with existing methods. (a) Self-Rewarding DPO \cite{yuan2024self} relies on a single reward criterion to rank sampled responses and selects both the highest-ranked (preferred) and lowest-ranked (rejected) responses for DPO training. (b) Structured Preference Optimization (ours) introduces a structured scoring framework with multiple criteria and an adaptive preference selection strategy, enabling more fine-grained and informed optimization.}
    \label{fig:compare}
    \vspace{-7mm}
\end{figure}

In autonomous systems, there is a growing demand for robots capable of executing complex, real-world tasks in domestic environments. Tasks such as organizing a room, preparing a meal, and cleaning up afterward require not only a diverse set of actions but also sophisticated long-term planning capabilities.
However, current approaches struggle with long-horizon tasks due to a lack of learning in long-term planning ability and the fact that most benchmarks \cite{puig2018virtualhome,liao2019synthesizing, shridhar2020alfred, shridhar2020alfworld} focus on short-term discrete tasks.
This gap hinders progress toward robots capable of handling the complex, multi-step tasks demanded by real-life scenarios.

Existing reasoning-based decision-making methods primarily rely on prompting strategies or environmental feedback to determine actions, often without explicitly modeling the quality of reasoning chains. While recent approaches \cite{yao2022react, zhao2024large, zhi2024pragmatic} leverage textual inputs for reasoning, they lack a structured mechanism to incorporate multimodal information or refine reasoning processes over extended horizons. Furthermore, prior optimization frameworks, such as Self-Rewarding DPO \cite{yuan2024self}, rely on a single reward criterion, which may lead to suboptimal preference selection as in Figure \ref{fig:compare}.

To address these limitations, we propose Structured Preference Optimization (SPO), a novel framework designed to enhance reasoning quality and decision-making in long-horizon task planning through structured preference evaluation and progressive learning. SPO consists of two core components:
1) \textbf{Structured Preference Evaluation and Optimization:} SPO systematically evaluates reasoning chains along two key dimensions---Textual Coherence, which assesses task relevance and historical consistency, and Image Awareness, which measures alignment with visual observations. These evaluations are used to construct high-quality preference pairs that explicitly guide the model toward superior reasoning steps, thereby enhancing decision reliability in complex multimodal tasks.
2) \textbf{Curriculum-Guided Progressive Learning:} SPO utilizes a progressive curriculum, incrementally increasing task complexity during training. This structured progression helps the model develop robust reasoning strategies and enhances generalization to diverse long-horizon scenarios, ensuring consistent real-world performance.

Finally, to bridge the notable gap in the field regarding the absence of a benchmark tailored for long-horizon tasks, we propose ExtendaBench, a comprehensive benchmark that categorizes the task into four difficulty levels based on the number of steps required for completion, namely ultra-short, short, medium, and long.
Leveraging the generative capabilities of GPT-4o \cite{openai2023gpt4o}, we create a diverse and extensive collection of tasks. These tasks undergo minimal human refinement to ensure high-quality data while significantly reducing the costs and effort associated with manual data labeling.

Our contributions can be summarized as follows:
\begin{itemize}[topsep=-2pt,leftmargin=8pt]
\setlength\itemsep{-0.3em}
    \item We introduce Structured Preference Optimization (SPO), a framework that enhances long-horizon reasoning through structured preference-based evaluation and curriculum-guided learning, enabling more effective decision-making.
    \item We propose ExtendaBench, a benchmark with four levels of difficulty and 1,509 tasks across VirtualHome and Habitat 2.0, providing a comprehensive evaluation suite for sustained reasoning in long-horizon task planning.
    \item We validate SPO through extensive experiments, demonstrating state-of-the-art performance in long-horizon task planning.
\end{itemize}

%% file: sec/2_related_work.tex
\section{Related Work}
\subsection{Multimodal Large Language Models}
The emergence of LLMs \cite{touvron2023llama, vicuna2023} has driven substantial progress in multimodal large language models (MLLMs), which aim to integrate both visual and textual modalities, advancing toward a more generalized form of intelligence. Early works such as BLIP-2 \cite{jian2024bootstrapping}, MiniGPT-4 \cite{zhu2023minigpt}, LLaVA \cite{liu2024visual}, and OpenFlamingo \cite{awadalla2023openflamingo} capitalized on pretrained vision encoders paired with LLMs, demonstrating strong performance in tasks like visual question answering and image captioning. mPLUG-Owl \cite{ye2023mplug} introduces a modularized training framework to further refine cross-modal interactions. On the closed-source side, models such as GPT-4V \cite{OpenAI2023GPT4TR} and Gemini \cite{team2023gemini}
pushes the boundaries of multimodal reasoning and interaction capabilities.
Unlike general-purpose MLLMs, we repurpose them for structured training in embodied planning tasks.

\subsection{LLM Self-improvement}
Self-improvement methods enhance LLMs by training on their own generated outputs.
These methods often involve supervised fine-tuning (SFT) on high-quality responses generated by the models themselves \cite{li2023self,wang2024self} or preference optimization \cite{yuan2024self,rosset2024direct,pang2024iterative,prasad2024self,zhang2024chain,jiang2024modality}, where the model is trained to distinguish between better and worse responses. These approaches mostly employ LLM-as-a-Judge prompting \cite{zheng2024judging} or train strong reward models \cite{xu2023some,havrilla2024glore} to evaluate and filter generated data, thereby guiding the model toward improved performance.
Unlike prior methods, we introduce structured preference optimization with targeted pair selection and curriculum learning for long-horizon embodied tasks.

\subsection{Embodied Task Planning}
Traditional robotics planning methods have relied on search algorithms in predefined domains \cite{fikes1971strips, garrett2020pddlstream, jiang2018task}, but face scalability challenges in complex environments with high branching factors \cite{puig2018virtualhome, shridhar2020alfred}. Heuristics have helped alleviate these limitations, leading to advancements \cite{baier2009heuristic, hoffmann2001ff, helmert2006fast, bryce2007tutorial}. More recently, learning-based methods like representation learning and hierarchical strategies have emerged, showing effectiveness in complex decision-making \cite{eysenbach2019search, xu2018neural, xu2019regression, srinivas2018universal, kurutach2018learning, nair2019hierarchical, jiang2019language}.
The advent of LLMs has further revolutionized planning by enabling task decomposition and robust reasoning \cite{li2022pre, huang2022inner, ahn2022can, valmeekam2022large, silver2022pddl, song2023llm, rana2023sayplan, driess2023palm, liu2023reflect, wu2023tidybot, wake2023chatgpt, chen2023llm, bhat2024grounding, zhi2024pragmatic}. Other works focus on translating natural language into executable code and formal specifications \cite{vemprala2023chatgpt, liang2023code, silver2023generalized, xie2023translating, skreta2023errors, liu2023llm+, zhang2023large, ding2023task,ding2023integrating, zhao2024large}. Some approaches fine-tune LLMs for better performance \cite{driess2023palm, qiu2023embodied}, while others opt for few-shot or zero-shot methods \cite{huang2022inner,huang2022language,singh2023progprompt} to avoid the resource demands of model training.
In contrast, our method introduces multimodal preference optimization, fine-grained preference scoring, and curriculum-guided optimization.

%% file: sec/3_method.tex
\begin{figure*}[t]
    \centering
    \includegraphics[width=\textwidth]{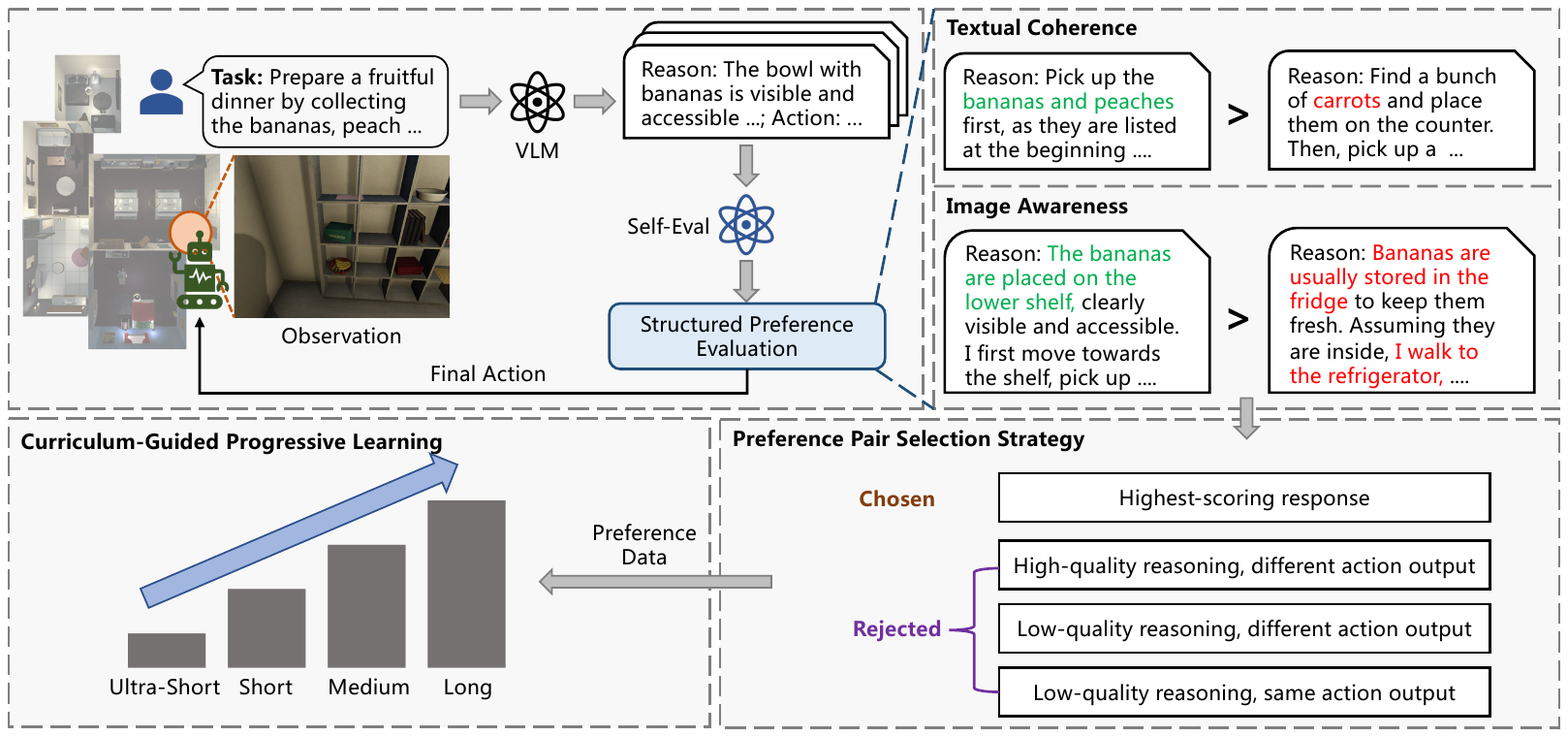}
    \vspace{-4mm}
    \caption{Overview of the Structured Preference Optimization. The SPO consists of three key components: 1) Structured Preference Evaluation, which systematically scores reasoning-action pairs based on textual coherence and image awareness; 2) Preference Pair Selection Strategy, which refines response optimization by selecting the highest-scoring reasoning-action pairs while rejecting low-quality alternatives based on structured criteria; 3) Curriculum-Guided Training, which progressively improves the model’s capabilities by adapting from ultra-short to long-horizon tasks through a staged optimization process.}
    \vspace{-6mm}
    \label{fig:method}
\end{figure*}

\section{Preliminaries}
Direct Preference Optimization (DPO) \cite{rafailov2024direct} is a reinforcement learning-free approach that optimizes a model’s policy using preference-labeled data. Instead of relying on an explicit reward model, DPO directly enforces preference ordering by encouraging the model to assign higher probabilities to preferred outputs over less preferred ones.

Given a dataset $D = \left\{ (x, y^+, y^-) \right\}$, where \(y^+\) is the preferred response and \(y^-\) is the less preferred response for input \(x\), DPO optimizes the following contrastive ranking loss:
\begin{align}
\mathcal{L}_{\text{DPO}}(\pi_\theta; \pi_{\text{ref}}) = - \mathbb{E}_{(x, y^+, y^-) \sim \mathcal{D}} \Bigg[ \log \sigma \Bigg( \notag \\
\beta \log \frac{\pi_\theta(y^+ \mid x)}{\pi_{\text{ref}}(y^+ \mid x)} 
- \beta \log \frac{\pi_\theta(y^- \mid x)}{\pi_{\text{ref}}(y^- \mid x)} 
\Bigg) \Bigg],
\label{eq:dpo}
\end{align}
where \(\sigma\) is the sigmoid function, and \(\beta\) is a scaling factor controlling preference sharpness.

\section{Structured Preference Optimization}
The Structured Preference Optimization (SPO) framework enhances long-horizon task planning by introducing a structured evaluation mechanism for reasoning quality and a progressive training strategy to improve model generalization. Unlike standard preference optimization, which lacks explicit reasoning quality assessment and task complexity adaptation, SPO systematically refines the model’s reasoning capabilities through Preference-Based Scoring and Optimization and Curriculum-Guided Training. The overview of our framework is shown in Figure \ref{fig:method}.

\subsection{Preference-Based Scoring and Optimization}
The structured preference-based optimization mechanism evaluates and ranks reasoning chains based on explicit criteria. Unlike standard preference optimization, which treats reasoning as a single scalar preference, SPO decomposes reasoning quality into multiple dimensions and optimizes the model’s decision-making accordingly.

\subsubsection{Structured Preference Evaluation}
\label{sec:preference_evaluation}
Instead of relying on external annotations, SPO adopts a self-evaluation approach, where the vision-language model (sLVLM) itself serves as the judge to assess reasoning quality. Given a generated reasoning chain \(R_i\), the model evaluates it based on the task context, which includes: task instruction ($I$), current image observation ($o$), and history of executed actions ($h$). Using this structured input, the model assigns two separate scores to assess different aspects of reasoning quality:
\begin{itemize}[topsep=-2pt,leftmargin=8pt]
\setlength\itemsep{-0.3em}
\item Textual Coherence ($S_{\text{text}}$): Evaluates the logical consistency of the reasoning chain, ensuring that each step is task-relevant and maintains historical consistency with prior steps. This prevents reasoning errors such as goal misalignment or contradictions in multi-step plans.
\item Image Awareness ($S_{\text{image}}$): Measures whether the reasoning chain sufficiently incorporates relevant information from the visual observations, ensuring that decisions are grounded in the environment rather than relying solely on textual priors.
\end{itemize}

To obtain these scores, the model is prompted with an evaluation query $p$, where the model $M$ estimates reasoning quality as follows:
\begin{align}
S_{\text{text}} &= M(p_{\text{text}}, R_i, I, h), \label{eq:text} \\
S_{\text{image}} &= M(p_{\text{image}}, R_i, I, o, h), \label{eq:image}
\end{align}
where \(p_{\text{text}}\) and \(p_{\text{image}}\) are evaluation prompts designed to assess textual coherence and image awareness, respectively. The overall preference score can then be computed as either a weighted combination:
\begin{equation}
    S(R_i) = w_1 \mathcal{S}_{\text{text}} + w_2 \mathcal{S}_{\text{image}},
    \label{equ:weighted_sum}
\end{equation}
where \(w_1\) and \(w_2\) are weighting factors that control the relative contribution of textual coherence and image awareness. Alternatively, instead of using a predefined weighted sum, the model directly provides an overall preference score:
\begin{equation}
    S(R_i) = M(p_{\text{overall}}, R_i, I, o, h),
    \label{equ:direct_scoring}
\end{equation}
where \(p_{\text{overall}}\) is an evaluation prompt requesting a single comprehensive score.
Empirically, we found that using the model to generate the overall preference score yields better optimization results compared to manually setting weighting factors.

\subsubsection{Preference Pair Selection Strategy}
To refine the model’s reasoning capabilities, SPO constructs structured preference pairs from model-generated samples, ensuring that the optimization process explicitly accounts for both reasoning quality and action selection. Unlike prior methods, which simply select the highest-scoring reasoning chain as the positive sample and the lowest-scoring reasoning chain as the negative sample, SPO introduces a targeted preference selection strategy that prevents the model from over-optimizing reasoning at the cost of decision accuracy.

Given a set of generated reasoning chains $\{ R_i \}$ for the same task input \((I, o, h)\), the model self-evaluates each reasoning chain using the scoring mechanism described in Structured Preference Evaluation.
The positive sample $R^+$ is selected as the highest-scoring reasoning chain, and in cases where multiple chains achieve the same highest score, we choose the one where the final action appears most frequently across all generated samples. This ensures that the model prioritizes common and stable action choices, reducing the risk of selecting an outlier action due to randomness in generation.
For the negative sample $R^-$, instead of always selecting the lowest-scoring reasoning chain, SPO considers different selection strategies to ensure both reasoning quality and action feasibility are optimized. The negative sample is chosen from one of the following categories:
\begin{itemize}[topsep=-2pt,leftmargin=8pt]
\setlength\itemsep{-0.3em}
\item High-quality reasoning, different action output: Reasoning chain with high preference scores but a different final action from $R^+$. This discourages optimizing reasoning quality while overlooking action correctness.
\item Low-quality reasoning, different action output: Reasoning chain with low preference scores and incorrect final action. This clarifies distinctions between poor reasoning and high-quality thought processes.
\item Low-quality reasoning, same action output: Reasoning chain with low preference scores but identical final action to $R^+$. This prevents the model from focusing solely on reasoning quality without validating decision consistency.
\end{itemize}

\subsubsection{Preference Optimization}
Once the structured preference pairs $(R^+, R^-)$ are selected, SPO directly applies DPO to align the model’s policy with the preferred reasoning chains. The optimization follows the original DPO contrastive ranking loss (referencing Eq. \ref{eq:dpo}, adapted to our task setting with inputs $(I, o, h)$:
\begin{align}
\mathcal{L}_{\text{pref}}(\pi_\theta; \pi_{\text{ref}}) 
&= - \mathbb{E}_{(I, o, h, R^+, R^-) \sim \mathcal{D}} 
\Big[ \log \sigma \Big( \notag \\
& \beta \log \frac{\pi_\theta(R^+ \mid I, o, h)}{\pi_{\text{ref}}(R^+ \mid I, o, h)} \notag \\
& - \beta \log \frac{\pi_\theta(R^- \mid I, o, h)}{\pi_{\text{ref}}(R^- \mid I, o, h)} 
\Big) \Big].
\label{eq:pref_loss}
\end{align}

\subsection{Curriculum-Guided Progressive Learning}
To facilitate structured learning, SPO categorizes tasks into four levels: ultra-short, short, medium, and long-horizon tasks. Instead of training on all task types simultaneously, SPO follows a progressive training strategy to gradually expose the model to increasing task complexity while preventing catastrophic forgetting.

Training is divided into four stages, where the model starts with ultra-short tasks and progressively incorporates more complex tasks in each subsequent stage. At every stage, a certain amount of previously learned tasks is retained to reinforce fundamental reasoning skills and prevent the model from overfitting to newly introduced tasks. This approach ensures that earlier-learned reasoning strategies remain effective as the model learns to handle longer task horizons.

A key challenge in curriculum learning is stabilizing the transition between different difficulty levels without disrupting previously learned decision patterns. To address this, SPO maintains a dynamic balance between newly introduced tasks and previously learned ones. During each training phase, the model is exposed to a mixture of current-stage tasks and replayed tasks from earlier stages, ensuring that it can generalize across task difficulties while refining long-horizon reasoning capabilities.

%% file: sec/4_benchmark.tex
\section{ExtendaBench}

\begin{figure*}[t]
    \centering
    \includegraphics[width=\textwidth]{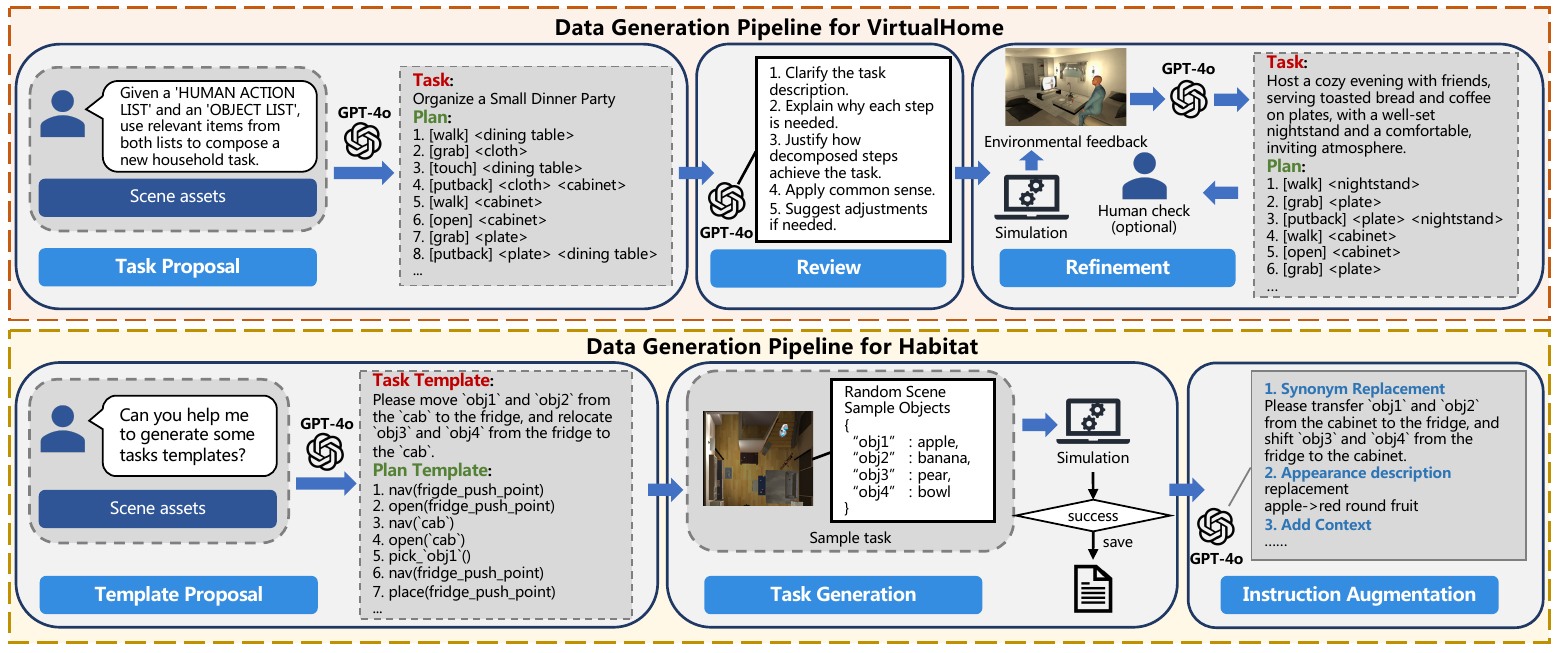}
    \vspace{-6mm}
    \caption{Data generation pipeline for ExtendaBench. (Top) In VirtualHome, GPT-4o generates tasks based on scene assets and a human action list, followed by plan generation, review, and refinement using simulation and feedback. (Bottom) In Habitat, scene objects are sampled and filled into task templates to create executable plans, which are validated in simulation. Instructions are then augmented with synonyms, appearance descriptions, and contextual cues to enhance linguistic diversity.}
    \vspace{-6mm}
    \label{fig:generate_task}
\end{figure*}

\begin{figure}[t]
    \centering
    \includegraphics[width=\linewidth]{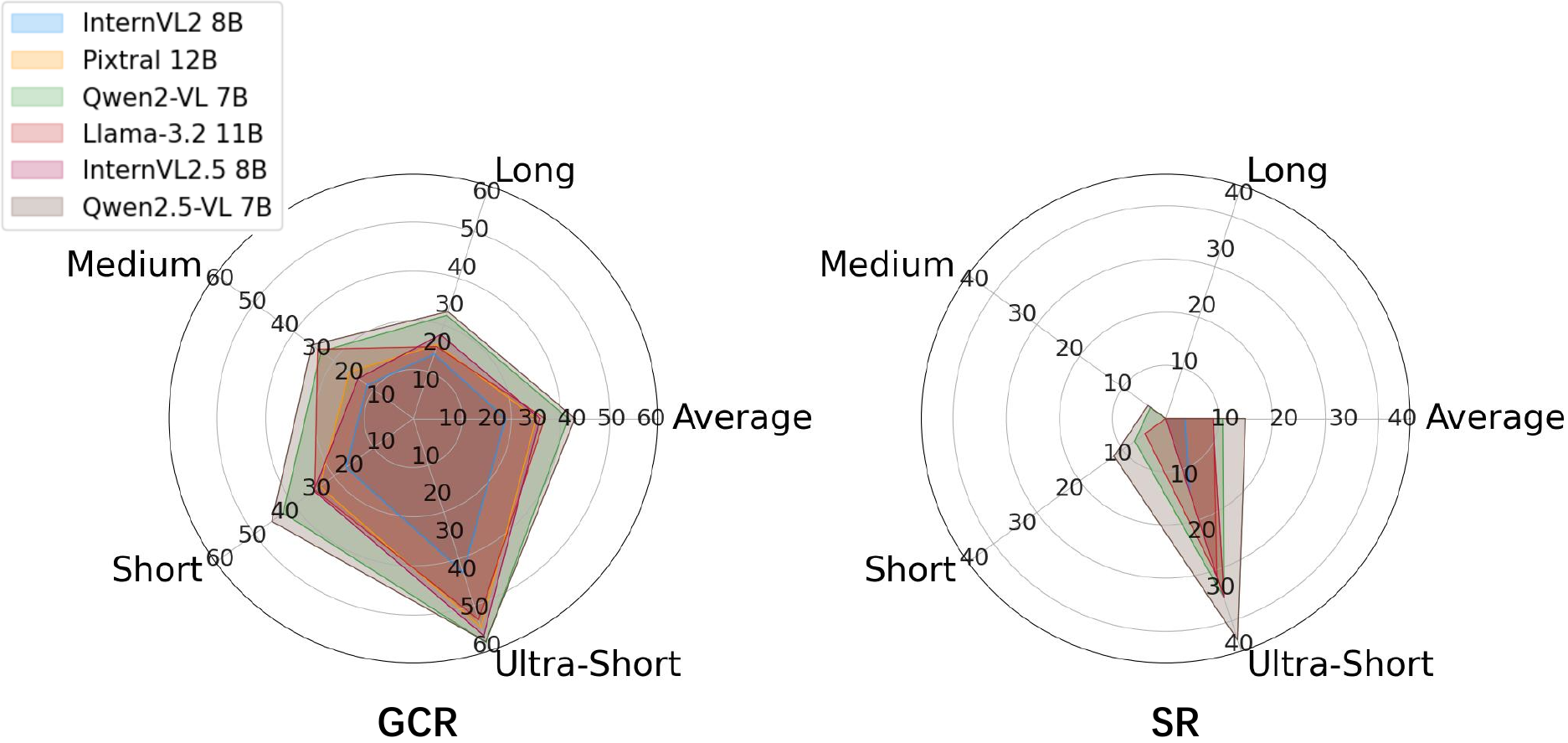}
    \vspace{-4mm}
    \caption{Comparison of various small vision-language models on different sets of our ExtendaBench in VirtualHome.}
    \vspace{-8mm}
    \label{fig:radar_llm}
\end{figure}

The ExtendaBench task corpus is developed using tailored approaches for each simulator, both leveraging GPT-4o’s advanced generative capabilities. For VirtualHome \cite{puig2018virtualhome}, we utilize GPT-4o to directly generate diverse and complex tasks, allowing for a wide range of scenarios. For Habitat 2.0 \cite{szot2021habitat}, GPT-4o is used to generate pre-defined templates as well as to create specific task instances from these templates, resulting in systematically varied tasks with extended action sequences that are suitable for long-horizon planning.

\subsection{VirtualHome}

\noindent
\textbf{Task Proposal}
The initial phase begins within the confines of VirtualHome, a simulated environment, where a varied collection of objects sets the stage for a multitude of task scenarios. By employing GPT-4o as a task generator, we design tasks focusing on object manipulation, striving for a wide array of task varieties and complexities. This method ensures an exhaustive representation of scenarios that closely mimic real-world challenges. To facilitate the generator's task creation, we provide prompts that are carefully constructed to inspire a broad range of tasks.

\noindent
\textbf{Review}
In the subsequent phase, GPT-4o undertakes the generation of detailed action plans for the devised tasks, meticulously outlining the steps required for successful task execution. To ensure the feasibility and coherence of these tasks, we introduce an additional examiner of scrutiny, also powered by GPT-4o. This examiner evaluates each task and its associated action plan for clarity, necessity, and coherence of steps, as well as the relevance and practicality of the actions and items involved, ensuring they belong to the simulated environment VirtualHome. It also assesses each step for common sense applicability, providing constructive feedback for further refinement.

\noindent
\textbf{Refinement}
After undergoing expert scrutiny, the generator refines the tasks and their corresponding action plans. Subsequent simulation of these revised tasks and plans enables further improvements based on simulator feedback. Tasks that are successfully executed within the simulator receive preliminary approval. Nevertheless, to guarantee optimal quality and applicability, we subject each task to a rigorous manual review, evaluating them for practicality and realism. Tasks that do not achieve success in the simulation are minimally modified by human according to the simulator's feedback, focusing on enhancing their realism and feasibility. 

The multi-stage process, with minimal human intervention, is designed to ensure the reliability and quality of the tasks and their associated plans.
The whole process of generating tasks in benchmark is shown in Figure \ref{fig:generate_task}.

\subsection{Habitat 2.0}
Building on the idea of Language Rearrangement \cite{szot2023large}, we replace its hand-crafted, short-horizon templates with an LLM-driven pipeline (Figure~\ref{fig:generate_task}) that automatically writes task schemas and expands them into markedly longer action sequences.

\noindent
\textbf{Template Proposal}
GPT-4o generates initial task templates based on scene assets. These templates define general task structures (e.g., moving objects between locations) and serve as the basis for generating varied instructions.

\noindent
\textbf{Task Generation}
Using the task templates, we sample objects within random scenes to generate specific tasks with extended action sequences. This phase results in more complex task plans that evaluate an agent's capacity for long-term planning and adaptability.

\noindent
\textbf{Instruction Augmentation}
To increase task diversity, we apply various transformations to the instructions. These include synonym replacement, appearance description alterations (e.g., “apple” to “red round fruit”), and additional contextual details. This augmentation, powered by GPT-4o, allows us to expand the instruction set, testing the agent's understanding and flexibility in interpreting varied language inputs.

\begin{table*}[t]
    \centering
    \caption{Comparison with existing methods using Qwen2.5-VL 7B as the baseline on different sets of our ExtendaBench in VirtualHome.}
    \vspace{-2mm}
    \small
    \resizebox{1.0\textwidth}{!}{
    \begin{tabular}{c|cc|cc|cc|cc|cc}
        \toprule
         & \multicolumn{2}{c|}{\bf \cellcolor[HTML]{F5E6E9} Ultra-Short} & \multicolumn{2}{c|}{\bf \cellcolor[HTML]{E2E6E1} Short} & \multicolumn{2}{c|}{\bf \cellcolor[HTML]{F9F2EB} Medium} & \multicolumn{2}{c|}{\bf \cellcolor[HTML]{D2F4F2} Long} & \multicolumn{2}{c}{\bf \cellcolor[HTML]{CDD4DF} Average} \\
        \midrule
        \bf Method & GCR & SR & GCR & SR & GCR & SR & GCR & SR & GCR & SR \\
        \midrule
        Baseline & 57.32 & 35.00 & 42.72 & 9.62 & 30.57 & 3.33 & 27.47 & 0 & 39.52 & 11.99 \\
        CoT \cite{wei2022chain} & 68.66 & 41.67 & 35.46 & 3.85 & 36.36 & 1.67 & 20.45 & 0 & 40.23 & 11.80 \\
        \midrule
        \multicolumn{1}{l|}{Self-Rewarding \cite{yuan2024self}} \\
        \cmidrule(lr){2-11}
        \textit{Iteration 1} & 62.13 & 35.00 & 42.89 & 7.69 & 36.38 & 3.33 & 22.55 & 0 & 40.99 & 11.51 \\
        \textit{Iteration 2} & 59.15 & 31.67 & 44.47 & 11.54 & 29.92 & 3.33 & 28.80 & 0 & 40.58 & 11.64 \\
        \textit{Iteration 3} & 58.93 & 35.00 & 48.16 & 9.62 & 32.56 & 3.33 & 27.46 & 0 & 41.78 & 11.99 \\
        \midrule
        \multicolumn{1}{l|}{Iterative RPO \cite{pang2024iterative}} \\
        \cmidrule(lr){2-11}
        \textit{Iteration 1} & 67.03 & 38.33 & 41.59 & 7.69 & 26.97 & 1.67 & 26.06 & 0 & 40.41 & 11.92 \\
        \textit{Iteration 2} & 72.31 & 43.33 & 40.06 & 1.92 & 31.66 & 3.33 & 22.33 & 0 & 41.73 & 12.15 \\
        \textit{Iteration 3} & 59.86 & 31.67 & 46.42 & 11.54 & 34.02 & 6.67 & 29.06 & 0 & 42.34 & 12.47 \\
        \midrule
        \rowcolor{lightblue}
        SPO (1 iteration) & 71.53 & 48.33 & 48.96 & 13.46 & 38.92 & 3.33 & 31.41 & 2.17 & \textbf{47.71} & \textbf{16.83} \\
        \bottomrule
    \end{tabular}}
    \vspace{-6mm}
    \label{tab:compare_vh}
\end{table*}

\subsection{Dataset Statistics}
The categorization within ExtendaBench is defined by the length of the action sequence required to accomplish a task, distributed as follows:
\begin{itemize}[topsep=-2pt,leftmargin=8pt]
\setlength\itemsep{-0.3em}
    \item Ultra-Short Tasks: Tasks that can be completed in fewer than 10 actions.
    \item Short Tasks: Tasks requiring 10 to 20 actions for completion.
    \item Medium Tasks: Tasks necessitating 20 to 30 actions to finish.
    \item Long Tasks: Tasks that demand more than 30 actions to complete.
\end{itemize}
The VirtualHome set includes a total of 605 tasks, with 220 ultra-short tasks, 128 short tasks, 155 medium tasks, and 102 long tasks. Similarly, the Habitat 2.0 set comprises 904 tasks, distributed as 161 ultra-short tasks, 243 short tasks, 190 medium tasks, and 310 long tasks.

%% file: sec/5_experiments.tex
\section{Experiments}

\begin{table*}[t]
    \centering
    \caption{Results using Qwen2.5-VL 7B as the baseline on different sets of our ExtendaBench in Habitat.}
    \vspace{-2mm}
    \small
    \resizebox{1.0\textwidth}{!}{
    \begin{tabular}{c|cc|cc|cc|cc|cc}
        \toprule
         & \multicolumn{2}{c|}{\bf \cellcolor[HTML]{F5E6E9} Ultra-Short} & \multicolumn{2}{c|}{\bf \cellcolor[HTML]{E2E6E1} Short} & \multicolumn{2}{c|}{\bf \cellcolor[HTML]{F9F2EB} Medium} & \multicolumn{2}{c|}{\bf \cellcolor[HTML]{D2F4F2} Long} & \multicolumn{2}{c}{\bf \cellcolor[HTML]{CDD4DF} Average} \\
        \midrule
        \bf Method & GCR & SR & GCR & SR & GCR & SR & GCR & SR & GCR & SR \\
        \midrule
        Baseline & 41.67 & 33.33 & 14.39 & 8.57 & 3.17 & 0 & 2.48 & 0 & 15.43 & 10.48 \\
        CoT \cite{wei2022chain} & 42.36 & 50.00 & 9.49 & 8.57 & 7.51 & 0 & 6.10 & 0 & 16.36 & 14.64 \\
        \midrule
        \multicolumn{1}{l|}{Self-Rewarding \cite{yuan2024self}} \\
        \cmidrule(lr){2-11}
        \textit{Iteration 1} & 43.06 & 36.11 & 13.33 & 8.57 & 3.32 & 0 & 2.48 & 0 & 15.55 & 11.17 \\
        \textit{Iteration 2} & 43.06 & 33.33 & 14.19 & 8.57 & 4.34 & 0 & 2.48 & 0 & 16.01 & 10.48 \\
        \textit{Iteration 3} & 44.44 & 36.11 & 13.59 & 8.57 & 3.63 & 0 & 2.48 & 0 & 16.03 & 11.17 \\
        \midrule
        \multicolumn{1}{l|}{Iterative RPO \cite{pang2024iterative}} \\
        \cmidrule(lr){2-11}
        \textit{Iteration 1} & 45.14 & 36.11 & 12.90 & 8.57 & 3.43 & 0 & 2.86 & 0 & 16.08 & 11.17 \\
        \textit{Iteration 2} & 33.33 & 38.89 & 12.17 & 11.43 & 11.35 & 0 & 7.62 & 0 & 16.12 & 12.58 \\
        \textit{Iteration 3} & 38.54 & 44.44 & 15.06 & 8.57 & 10.49 & 0 & 6.76 & 0 & 17.71 & 13.25 \\
        \midrule
        \rowcolor{lightblue}
        SPO (1 iteration) & 52.08 & 50.00 & 14.78 & 11.43 & 10.51 & 0 & 6.67 & 0 & \textbf{21.01} & \textbf{15.36} \\
        \bottomrule
    \end{tabular}}
    \vspace{-2mm}
    \label{tab:compare_habitat}
\end{table*}

\begin{table*}[t]
    \centering
    \caption{Ablation studies of different modules in VirtualHome.}
    \vspace{-2mm}
    \small
    \resizebox{1.0\textwidth}{!}{
    \begin{tabular}{ccc|cc|cc|cc|cc|cc}
        \toprule
         & & & \multicolumn{2}{c|}{\bf \cellcolor[HTML]{F5E6E9} Ultra-Short} & \multicolumn{2}{c|}{\bf \cellcolor[HTML]{E2E6E1} Short} & \multicolumn{2}{c|}{\bf \cellcolor[HTML]{F9F2EB} Medium} & \multicolumn{2}{c|}{\bf \cellcolor[HTML]{D2F4F2} Long} & \multicolumn{2}{c}{\bf \cellcolor[HTML]{CDD4DF} Average} \\
        \midrule
        \bf Textual & \bf Image & \bf Curriculum & GCR & SR & GCR & SR & GCR & SR & GCR & SR & GCR & SR \\
        \midrule
        \xmark & \xmark & \xmark & 67.25 & 36.67 & 34.50 & 1.92 & 34.58 & 3.33 & 23.09 & 0 & 39.86 & 10.48 \\
        \cmark & \xmark & \xmark & 71.70 & 43.33 & 45.52 & 9.62 & 30.57 & 1.67 & 16.71 & 0 & 41.13 & 13.65 \\
        \xmark & \cmark & \xmark & 69.71 & 40.00 & 42.11 & 1.92 & 25.98 & 3.33 & 26.16 & 0 & 40.99 & 11.31 \\
        \cmark & \cmark & \xmark & 70.52 & 43.33 & 43.89 & 7.69 & 33.96 & 3.33 & 27.90 & 0 & 44.07 & 13.59 \\
        \cmark & \cmark & \cmark & 71.53 & 48.33 & 48.96 & 13.46 & 38.92 & 3.33 & 31.41 & 2.17 & \textbf{47.71} & \textbf{16.83} \\
        \bottomrule
    \end{tabular}}
    \vspace{-6mm}
    \label{tab:ablation_study}
\end{table*}

\subsection{Experimental Setup}
For the VirtualHome set, we designate 218 tasks as the test set, with the remaining tasks serving as the training set. The Habitat 2.0 set also includes 120 test tasks. As our approach is unsupervised, we do not utilize the training set data for model training.

\noindent
\textbf{Evaluation Metrics}
To assess system efficacy, we employ success rate (SR) and goal conditions recall (GCR) \cite{singh2023progprompt} as our primary metrics. SR measures the proportion of executions where all key goal conditions (changing from the beginning to the end during a demonstration) are satisfied. GCR calculates the discrepancy between the expected and achieved end state conditions, relative to the total number of specific goal conditions needed for a task. A perfect SR score of 100\% corresponds to achieving a GCR of 100\%.

\subsection{Comparison of Vision-Language Models}
\label{sec:compare_vlm}
To assess the capabilities of various small-scale vision-language models (sLVLMs) on long-horizon task planning, we evaluated six models---InternVL2 8B \cite{chen2024far}, Pixtral 12B \cite{agrawal2024pixtral}, Qwen2-VL 7B \cite{wang2024qwen2}, Llama-3.2 11B \cite{dubey2024llama}, InternVL2.5 8B \cite{chen2024expanding}, and Qwen2.5-VL 7B \cite{Qwen2.5-VL}---on our ExtendaBench benchmark in VirtualHome, spanning ultra-short to long tasks. Figure \ref{fig:radar_llm} presents the comparative performance in terms of GCR and SR across different task horizons. The results indicate that while all models perform well on ultra-short tasks, performance drops sharply as task complexity increases, with SR reaching 0\% on long tasks for most models. Among them, Qwen2.5-VL 7B achieves the highest average GCR and SR, demonstrating the best overall performance in long-horizon task planning.

\subsection{Comparison with Existing Methods}
We compare SPO with existing long-horizon reasoning methods, including Chain-of-Thought (CoT) \cite{wei2022chain}, as well as the multimodal VLM-based versions of Self-Rewarding \cite{yuan2024self} and Iterative RPO \cite{pang2024iterative}, using Qwen2.5-VL 7B \cite{Qwen2.5-VL} as the baseline. The evaluations are conducted on ExtendaBench in VirtualHome (Table \ref{tab:compare_vh}) and Habitat (Table \ref{tab:compare_habitat}), covering tasks of increasing complexity from Ultra-Short to Long.

ExtendaBench provides a structured evaluation protocol with graded task difficulty, enabling fine-grained analysis of model reasoning under increasing complexity. As shown in Tables~\ref{tab:compare_vh} and~\ref{tab:compare_habitat}, model performance consistently degrades on harder tasks, validating the benchmark’s ability to reveal reasoning limitations across different methods.

Across both benchmarks, CoT improves over the baseline in Habitat, particularly on shorter tasks, highlighting the benefits of explicit reasoning in simpler settings. However, its performance drops notably on longer tasks, where it lacks structured planning capabilities. In VirtualHome, CoT offers limited gains and struggles with complex scenarios.

Self-Rewarding and Iterative RPO introduce iterative refinement, leading to moderate improvements on short and medium tasks. However, both methods fail to generalize to long-horizon planning, with SR dropping to 0\% in long tasks across environments, indicating difficulties in maintaining coherent reasoning over extended sequences.

In contrast, SPO achieves the best overall performance in both VirtualHome and Habitat, outperforming all baselines. Notably, SPO delivers strong long-horizon reasoning without relying on iterative generation, achieving higher GCR and SR than Iterative RPO and Self-Rewarding across all difficulty levels.

\subsection{Ablation Study}
To evaluate the contributions of different components in SPO, we conduct an ablation study on VirtualHome, selectively removing textual coherence scoring, image awareness scoring, and curriculum-guided training. The results in Table \ref{tab:ablation_study} show that removing textual coherence scoring leads to the most significant performance drop, especially on short, medium, and long tasks, indicating its critical role in maintaining reasoning consistency. Removing image awareness scoring also results in a decline, particularly on long tasks, where integrating visual observations becomes more important. Without curriculum learning, performance on medium and long tasks deteriorates, demonstrating that progressive training helps the model handle more complex task sequences. The full SPO model achieves the highest performance, with 47.71\% GCR and 16.83\% SR, confirming that structured preference learning and curriculum-guided training together enable more effective long-horizon task planning.

%% file: sec/6_conclusion.tex
\section{Conclusion}
We introduce Structured Preference Optimization (SPO), a method for improving long-horizon vision-language task planning through structured preference learning and curriculum-guided training. Unlike existing methods that struggle with multi-step decision-making, SPO systematically evaluates reasoning chains based on textual coherence and image awareness, ensuring high-quality reasoning and action selection. Additionally, curriculum-guided training progressively adapts the model from simpler to more complex tasks, enhancing generalization and robustness in long-horizon scenarios.
To support research in this area, ExtendaBench provides a benchmark spanning VirtualHome and Habitat simulators with tasks of increasing difficulty. Experimental results show that SPO outperforms prior methods, particularly in long-horizon task planning, demonstrating improved reasoning consistency and decision-making accuracy.

\section*{Limitations}
While our proposed Structured Preference Optimization (SPO) framework demonstrates strong performance in long-horizon task planning, it is currently implemented using smaller-scale vision-language models to enable efficient training and extensive experimentation. This design choice allows for faster iteration and detailed analysis but may not fully reflect the potential of SPO when applied to larger, more capable models. Extending the framework to larger-scale models remains an important direction for future work, as it could further enhance reasoning ability and task performance in complex embodied environments.

\section*{Acknowledgements}
This work is supported by National Key Research and Development Program of China (2024YFE0203100), Scientific Research Innovation Capability Support Project for Young Faculty (No.ZYGXQNJSKYCXNLZCXM-I28), National Natural Science Foundation of China (NSFC) under Grants No.62476293,  Nansha Key R\&D Program under Grant No.2022ZD014, and General Embodied AI Center of Sun Yat-sen University.

%% file: sec/X_suppl.tex
\section{More Details for ExtendaBench}

\subsection{Statistics}

\subsubsection{Overview}
Table \ref{tab:data_type} provides a summary of key characteristics of the VirtualHome and Habitat datasets in our ExtendaBench, highlighting differences in scene complexity, task variety, and action requirements. The VirtualHome dataset consists of 7 distinct scenes with a total of 390 objects, supporting 294 task types across 605 instructions. In VirtualHome, the simulator provides 16 unique executable actions, enabling a broader range of task interactions.
In contrast, the Habitat dataset features 105 scenes with 82 distinct objects, enabling 20 task types across 904 instructions. The Habitat simulator supports 6 unique executable actions.

\begin{table}[h]
    \centering
    \caption{Overview of scene and task characteristics in VirtualHome and Habitat.}
    \begin{tabular}{c|c|c}
        \toprule
         & VirtualHome & Habitat \\
        \midrule
        Scene Number & 7 & 105 \\
        Scene Objects & 390 & 82 \\
        Task Type & 294 & 20 \\
        Instructions & 605 & 904 \\
        Action Number & 16 & 6 \\
        \bottomrule
    \end{tabular}
    \label{tab:data_type}
\end{table}

\subsubsection{Data Distribution Across Sets}
\textbf{VirtualHome}
For the VirtualHome dataset, tasks are categorized into ultra short, short, medium, and long. Each category includes a portion reserved for testing, with the remaining used for training. The distribution is as follows:
\begin{itemize}
    \item Ultra short: This category contains 220 tasks in total, with 46 allocated for testing and 174 for training.
    \item Short: A total of 128 tasks, with 60 reserved for testing and 68 for training.
    \item Medium: Comprising 155 tasks, with 52 for testing and 103 for training.
    \item Long: The most complex category, including 102 tasks in total, with 60 allocated for testing and 42 for training.
\end{itemize}

\noindent
\textbf{Habitat}
For Habitat, the dataset is similarly divided into four categories based on task length: ultra short, short, medium, and long. For each category, a portion of the tasks is allocated for testing, and the remaining are used for training. The details are as follows:
\begin{itemize}
    \item Ultra short: This category contains 161 tasks, with 36 reserved for testing and 125 for training.
    \item Short: There are 243 tasks, of which 35 are for testing and 208 for training.
    \item Medium: A total of 190 tasks, including 31 for testing and 159 for training.
    \item Long: The largest category, comprising 310 tasks, with 30 allocated for testing and 280 for training.
\end{itemize}

\subsubsection{Word Frequency Distribution}
\begin{figure}[h]
    \centering
    \includegraphics[width=\linewidth]{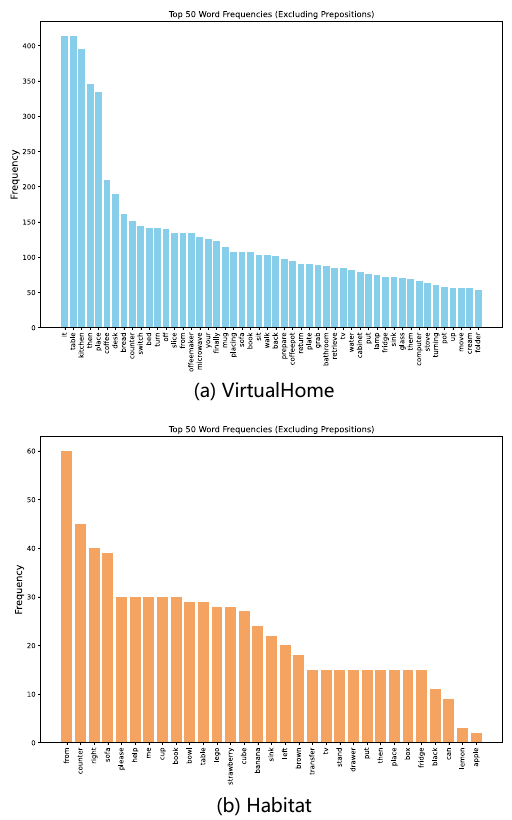}
    \caption{Word frequency analysis for ExtendaBench.}
    \label{fig:vocab}
\end{figure}

Figure \ref{fig:vocab} presents the top 50 most frequent words, excluding prepositions, in the datasets generated for VirtualHome and Habitat environments. Subfigure (a) shows the word frequencies from VirtualHome, highlighting terms associated with common objects and actions, such as ``table,'' ``kitchen,'' and ``place,'' reflecting its simulation of domestic scenarios. Subfigure (b) illustrates the word frequencies for Habitat, where terms like ``from,'' ``counter,'' and ``cup'' dominate, indicating tasks involving object interaction and spatial relationships.

\subsubsection{Action Lengths}
\begin{figure}[h]
    \centering
    \includegraphics[width=\linewidth]{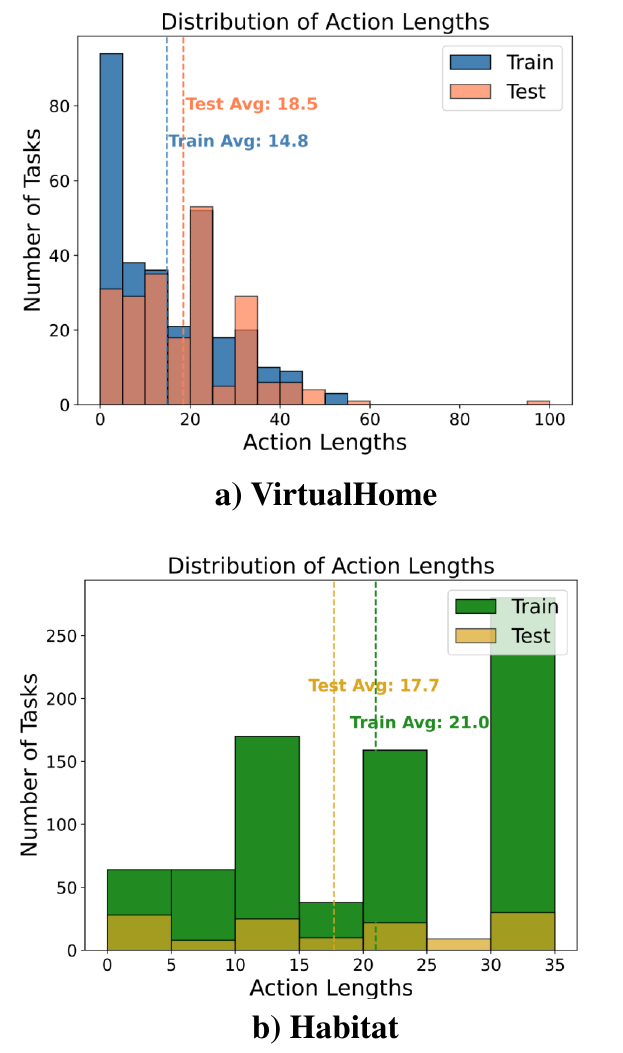}
    \caption{Distribution of action lengths in our benchmark.}
    \label{fig:action_len}
\end{figure}

Figure \ref{fig:action_len} splots primitive-action lengths and underscores the benchmark’s long-horizon nature: in VirtualHome the training split already ranges broadly (mean 14.8 actions) with a heavy tail extending to 58 steps, while an extreme task is held out for testing to enforce horizon extrapolation; Habitat pushes lengths even higher---training tasks centre around 20–35 actions (mean 21.0) and the test split, though slightly shorter on average (17.7), still requires multi-dozen-step plans---so across both environments the majority of tasks demand extended, sequential reasoning, and all subsequent results are reported per environment and split to reveal model performance along this length-generalisation axis.

\subsection{Task Complexity}
\subsubsection{Correlation between Action Length and Task Complexity}
Longer action chains inevitably accumulate execution error: under the all-or-nothing success metric, a single early slip invalidates the entire trajectory, so the chance of finishing a plan falls sharply as its length grows. In the VirtualHome portion of ExtendaBench, the split-wise statistics as in Table~\ref{tab:complexity} confirm this effect: average plan length stretches from 7.48 to 36.11 steps, instruction length from 18.1 to 61.7 tokens, and the syntactic-complexity score from 0.21 to 0.29, indicating deeper clause nesting and a larger set of entities that must be tracked. Empirically, the same Qwen2.5-VL-7B baseline that achieves 42\% GCR (33\% SR) on ultra-short VirtualHome tasks manages only 2\% GCR (0\% SR) on long ones, underscoring how these structurally richer instructions translate into far tougher planning problems. Taken together---error-accumulation theory, the monotonic rise in structural and linguistic complexity, and the observed performance cliff---demonstrate that action length is a well-grounded proxy for task difficulty within ExtendaBench.

\begin{table*}[h]
    \centering
    \caption{Correlation between action length and linguistic complexity across task difficulty levels in the VirtualHome subset of ExtendaBench.}
    \begin{tabular}{c|c|c|c}
        \toprule
         & Avg Action Length$\uparrow$ & Avg Instruction Length$\uparrow$& Avg Syntax Score~\cite{lu2010automatic}$\uparrow$ \\
        \midrule
        Ultra-Short & 7.48 &18.14 & 0.21 \\
        Short & 12.97 & 28.84 & 0.26 \\
        Medium & 22.46 & 43.96& 0.28 \\
        Long & 36.11 & 61.71& 0.29 \\
        \bottomrule
    \end{tabular}
    \label{tab:complexity}
\end{table*}

\subsubsection{Comparison of Task Complexity across Datasets}
ExtendaBench demonstrates significant advantages across multiple key metrics as in Table~\ref{tab:instruction-complexity}. Its average action length (18.60) substantially exceeds that of other datasets (ALFRED: 6.71, LLarp: 4.40, VirtualHome: 9.83), with the long-horizon subset reaching an average of 35.56 steps---highlighting its unique value for long-horizon reasoning tasks. Additionally, ExtendaBench exhibits the highest instruction length (31.15 tokens on average) and syntactic complexity score (0.31) among all compared datasets, indicating that its task instructions are both structurally complex and semantically rich.

\begin{table*}[ht]
\centering
\caption{Comparison of action, instruction, and language complexity across datasets.}
\resizebox{1.0\textwidth}{!}{
\begin{tabular}{lccc}
\toprule
Dataset &  Avg Action Length$\uparrow$ & Avg Instruction Length$\uparrow$& Avg Syntax Score~\cite{lu2010automatic}$\uparrow$ \\
\midrule
ALFRED~\cite{shridhar2020alfred}          & 6.71 & 9.26 &0.19 \\
LLarp~\cite{szot2023large}           & 4.40 &  10.59& 0.20 \\
VirtualHome~\cite{puig2018virtualhome}     & 9.83 & 19.11& 0.18 \\
\textbf{ExtendaBench (Ours)} & \textbf{18.60} &  \textbf{31.15}& \textbf{0.31} \\
\bottomrule
\end{tabular}}
\label{tab:instruction-complexity}
\end{table*}

\section{More Details for Experiments}
\label{sec:appendix_exp}

\subsection{Experimental Setup}
For data generation, we produce \(K = 5\) responses per prompt, employing a sampling temperature of 0.7 and a top-p value of 0.95. The generated dataset is then used to train the model for 3 epochs. During training, the learning rate is set to \( 2\mathrm{e}{-5} \). For LoRA, we use a rank value of 16, an alpha parameter of 32, and a dropout rate of 0.05. Unlike the Self-Rewarding framework, which involves iterative training where the trained model is used to re-label data and retrain in a loop, our approach trains the model only once, simplifying the training process while maintaining effectiveness.

\subsection{Comparison of Vision-Language Models}

In addition to the models discussed in Section 6.2, we also compare InternVL3 8B \cite{zhu2025internvl3} to assess whether Qwen2.5-VL 7B remains competitive despite having fewer parameters.
As shown in Table \ref{tab:compare_internvl3}, Qwen2.5-VL 7B achieves the highest average SR on both VirtualHome (11.99\% vs. 11.28\%) and Habitat (10.48\% vs. 9.78\%), and matches or outperforms InternVL3-8B on 11 of 16 split-metric pairs. While InternVL3-8B shows notable gains on medium/long tasks, Qwen2.5-VL 7B demonstrates more stable performance across environments and task lengths, confirming its reliability as a backbone model.

\begin{table*}[h]
    \centering
    \caption{Performance comparison between Qwen2.5-VL 7B and InternVL3 8B.}
    \vspace{-2mm}
    \resizebox{1.0\textwidth}{!}{
    \begin{tabular}{c|c|cc|cc|cc|cc|cc}
        \toprule
         & & \multicolumn{2}{c|}{Ultra-Short} & \multicolumn{2}{c|}{Short} & \multicolumn{2}{c|}{Medium} & \multicolumn{2}{c|}{Long} & \multicolumn{2}{c}{Average} \\
        \midrule
         & & GCR & SR & GCR & SR & GCR & SR & GCR & SR & GCR & SR \\
        \midrule
         \multirow{2}{*}{VirtualHome} & Qwen2.5-VL 7B & 57.32 & 35.00 & 42.72 & 9.62 & 30.57 & 3.33 & 27.47 & 0 & 39.52 & 11.99 \\
          & InternVL3 8B & 65.83 & 33.33 & 41.94 & 5.77 & 33.26 & 4.35 & 29.20 & 1.67 & 42.56 & 11.28 \\
          \midrule
         \multirow{2}{*}{Habitat} &Qwen2.5-VL 7B & 41.67 & 33.33 & 14.39 & 8.57 & 3.17 & 0 & 2.48 & 0 & 15.43 & 10.48 \\
          & InternVL3 8B & 33.68 & 30.56 & 12.36 & 8.57 & 6.23 & 0 & 2.48 & 0 & 13.69 & 9.78 \\
        \bottomrule
    \end{tabular}}
    \label{tab:compare_internvl3}
\end{table*}

\subsection{Effect of Preference Pair Selection}
To assess the impact of our preference pair selection strategy, we perform an ablation study using the Textual Coherence model (corresponding to row 2 in Table~\ref{tab:ablation_study}). As shown in Table~\ref{tab:select_pair}, enabling pair selection improves GCR from 40.04\% to 41.13\% and SR from 11.73\% to 13.65\%, yielding gains of +1.09 and +1.92 percentage points, respectively. These results indicate that structured preference selection contributes to more accurate decision-making by guiding the model with more informative comparisons.

\begin{table}[h]
    \centering
    \caption{Average performance of preference pair selection strategy in VirtualHome.}
    \begin{tabular}{c|cc}
        \toprule
        pair selection & GCR & SR \\
        \midrule
        \xmark & 40.04 & 11.73 \\
        \cmark & 41.13 & 13.65 \\
        \bottomrule
    \end{tabular}
    \label{tab:select_pair}
\end{table}

\subsection{Impact of Evaluator Strength}
To quantify how the capacity of the external evaluator influences learning, we replaced the default Qwen2.5-VL 7B assessor with GPT-4o and retrained under otherwise identical settings. As reported in Table \ref{tab:evaluators}, the stronger evaluator yields consistent gains on the VirtualHome benchmark, improving GCR from 47.71\% to 49.62\% and SR from 16.83\% to 19.26\%. These results confirm that higher-quality evaluators provide more informative preference signals, which in turn translate into better long-horizon task performance.

\begin{table}[h]
    \centering
    \caption{Compare with methods using different evaluators  in  VirtualHome environment.}
    \begin{tabular}{c|cc}
        \toprule
        evaluator & GCR & SR \\
        \midrule
        Qwen2.5-VL 7B & 47.71 & 16.83 \\
        GPT-4o & 49.62 & 19.26 \\
        \bottomrule
    \end{tabular}
    \label{tab:evaluators}
\end{table}

\subsection{Score Combination Strategies}
As described in Section \ref{sec:preference_evaluation}, we explore two approaches for combining the textual coherence score ($S_{\text{text}}$) and image awareness score ($S_{\text{image}}$). The first, shown in Equation \ref{equ:weighted_sum}, uses a weighted sum with tunable weights ($w_1$, $w_2$). The second, described in Equation \ref{equ:direct_scoring}, adopts a direct scoring approach, where the model is guided to first assess task alignment and image utilization independently, and then produce an overall score that integrates both aspects, following the prompt described in Section~\ref{sec:prompt_preference_evaluation}. This approach avoids manual weighting and achieves better empirical performance.
Table~\ref{tab:diff_combination} compares the two approaches, showing that direct scoring achieves the best performance among the tested settings and does not require manual tuning of combination weights.
We hypothesize two main reasons:
\begin{itemize}
    \item In a weighted sum, a chain that writes a very clear rationale can still receive a good overall score even if it references an object that does not exist in the image---the high text score masks the low image score. The direct-scoring judge, by contrast, looks at text and image evidence together and gives a high mark only when the reasoning is both logically consistent and visually grounded. This tighter evaluation criterion may explain the performance improvements reported for direct scoring.
    \item Optimal weights shift with task composition; small changes in $w_1, w_2$ can flip the ranking (SR spans from 9.17\% to 13.88\%). Direct scoring removes this hyper-parameter altogether, providing a stable criterion that generalises across lengths and environments.
\end{itemize}

\begin{table}[h]
    \centering
    \caption{Compare with methods using different combinations in the VirtualHome environment.}
    \begin{tabular}{cc|cc}
        \toprule
         & & GCR & SR \\
        \midrule
        \multicolumn{2}{c|}{weighted sum} & \\
        \cmidrule(lr){3-4}
        $w_1$=1.0 & $w_2$=1.0 & 41.66 & 9.17 \\
        $w_1$=1.0 & $w_2$=0.8 & 43.13 & 11.73 \\
        $w_1$=1.0 & $w_2$=0.5 & 41.59 & 13.46 \\
        $w_1$=0.8 & $w_2$=1.0 & 45.25 & 13.88 \\
        $w_1$=0.5 & $w_2$=1.0 & 41.81 & 11.99 \\
        \midrule
        \multicolumn{2}{c|}{direct scoring} & 47.71 & 16.83 \\
        \bottomrule
    \end{tabular}
    \label{tab:diff_combination}
\end{table}

\subsection{Effectiveness of Curriculum Learning}
Our curriculum consists of four sequential stages aligned with progressively increasing task complexity. To quantify the contribution of each curriculum stage, we evaluate model performance cumulatively after completing each stage, with each stage initialized from the checkpoint obtained in the previous one. Results in Table~\ref{tab:diff_curricuclum} show consistent improvements from Stage~1 through Stage~4, with GCR rising from 41.88\% to 47.71\% and SR improving from 11.80\% to 16.83\%. These gains highlight the effectiveness of our curriculum strategy in systematically enhancing model capabilities on complex long-horizon tasks.

\begin{table}[h]
    \centering
    \caption{Comparison of different stages in curriculum learning in the VirtualHome environment.}
    \begin{tabular}{c|cc}
        \toprule
         & GCR & SR \\
        \midrule
        stage 1 & 41.88 & 11.80 \\
        stage 2 & 43.21 & 14.78 \\
        stage 3 & 45.98 & 14.90 \\
        stage 4 & 47.71 & 16.83 \\
        \bottomrule
    \end{tabular}
    \label{tab:diff_curricuclum}
\end{table}

\subsection{Comparison between SPO and Single-negative DPO}
We ran standard single-negative DPO with each of our three negative-pair rules in isolation and compared them to SPO, which retains all three rules simultaneously. The results in Table~\ref{tab:single_dpo} show that every single-rule variant improves over the vanilla model to a similar extent (e.g., “low-quality reasoning, same action” reaches 44.88 GCR / 15.32 SR), yet none matches the full SPO configuration: combining the three complementary negatives pushes performance to 47.71 GCR / 16.83 SR, a further gain of +2.8 GCR and +1.5 SR over the best single-rule baseline. This demonstrates that the additional negative categories capture distinct failure modes and that their union provides the strongest learning signal, clarifying the specific benefit of our multi-type sampling strategy over conventional single-negative DPO.

\begin{table}[h]
    \centering
    \caption{Comparison between SPO and standard single-negative DPO in VirtualHome.}
    \resizebox{1.0\linewidth}{!}{
    \begin{tabular}{c|cc}
        \toprule
         & GCR & SR \\
        \midrule
        High-quality reasoning, different action & 44.52 & 14.01 \\
        Low-quality reasoning, different action & 44.12 & 14.07 \\
        Low-quality reasoning, same action & 44.88 & 15.32 \\
        SPO (three rules together) & 47.71 & 16.83 \\
        \bottomrule
    \end{tabular}}
    \label{tab:single_dpo}
\end{table}

\subsection{Comparison with Existing LLM-based Methods}
Our task formulation assumes that the agent receives only image observations and textual instructions as input, requiring it to infer actions solely from visual context without relying on externally provided ground-truth object identities or positions. In contrast, prior frameworks such as SayCan~\cite{brohan2023can} and ProgPrompt~\cite{singh2023progprompt} depend on explicit object-level annotations, making them less suited to realistic embodied environments. To facilitate meaningful comparison, we adapted both methods by restricting inputs to task instructions and raw visual observations, removing access to ground-truth environment information. Table~\ref{tab:saycan} shows that under these consistent input constraints, our approach significantly outperforms SayCan and ProgPrompt on the VirtualHome benchmark.

The observed performance gap arises due to inherent limitations of these approaches when restricted to our realistic input scenario. SayCan, which relies on scoring all potential atomic actions with a language model, struggles with the combinatorial explosion of action-object pairs (over 2,000 possibilities) in our setting, causing inefficiencies and unreliable scoring---particularly with smaller models. ProgPrompt also faces challenges: firstly, it was originally designed to leverage large language models such as GPT3, whereas our setup employs a smaller vision-language model with constrained structured reasoning and code generation capabilities; secondly, smaller models inherently struggle with long-context comprehension, hindering their ability to generate coherent, visually-grounded multi-step programs aligned with historical context.

\begin{table}[h]
    \centering
    \caption{Comparison with SayCan and ProgPrompt in the VirtualHome environment.}
    \resizebox{1.0\linewidth}{!}{
    \begin{tabular}{c|cc}
        \toprule
         & GCR & SR \\
        \midrule
        SayCan \cite{brohan2023can} & 40.80 & 10.83 \\
        ProgPrompt \cite{singh2023progprompt} & 39.64 & 9.17\\
        Ours & 47.71 & 16.83\\
        \bottomrule
    \end{tabular}}
    \label{tab:saycan}
\end{table}

\subsection{Details for High-Quality Reasoning Selection}
For each task instance, all reasoning chains are first ranked by their overall scores. We identify the subset of chains that achieve the highest score. If multiple top-scoring responses result in different final actions, we select the one with the most frequently occurring action as the preferred output. The remaining top-scoring variants are treated as high-quality reasoning samples with differing actions.

In cases where only a single top-scoring response exists, we also include reasoning chains whose scores fall within a margin of 0.1 points from the maximum and lead to different final actions. These are likewise categorized as high-quality reasoning but are used as negative examples during training, as their final actions deviate from the preferred one. This distinction encourages the model to differentiate between logically coherent reasoning and correct decision-making.

This combination of threshold-based filtering and action consistency selection helps reduce the bias in self-assessment and prevents the model from over-optimizing for reasoning fluency alone.

\subsection{SPO vs. GRPO with Structured Preference Rewards}
To keep training cost tractable we ran all GRPO~\cite{shao2024deepseekmath} baselines with the smaller Qwen2.5-VL 3B backbone on Habitat (as shown in the Table~\ref{tab:grpo}). Our SPO (DPO-style) already surpasses vanilla GRPO (+4.38 pp GCR, +1.41 pp SR). When GRPO is augmented with the same Structured Preference rewards, its score rises further to 20.18 / 16.07, narrowing the margin to SPO to $\leq$ 1.4 pp. This result shows (i) Structured Preference signals are the key performance driver---whichever optimisation is used---and (ii) our DPO-based SPO achieves comparable performance with a simpler, resource-efficient training loop.

\begin{table}[h]
    \centering
    \caption{Compare with GRPO-based methods using Qwen2.5-VL 3B as the base model in Habitat environment.}
    \resizebox{1.0\linewidth}{!}{
    \begin{tabular}{c|cc}
        \toprule
         & GCR & SR \\
        \midrule
        baseline & 12.91 & 11.17 \\
        SPO & 19.83 & 14.66 \\
        GRPO & 15.45 & 13.25 \\
        GRPO w/ Structured Preference rewards & 20.18 & 16.07 \\
        \bottomrule
    \end{tabular}}
    \label{tab:grpo}
\end{table}

\subsection{Results on other Planning Datasets}
We centred our study on ExtendaBench because, unlike prior embodied-planning benchmarks, it spans multiple difficulty tiers---ultra-short to long (up to 60 primitive actions) within a single RGB-grounded suite. This breadth of plan complexity is exactly the setting for which Structured Preference Optimisation (SPO) was designed. To give an additional reference point, we ran a quick check on ALFRED: using only 10\% of the ALFRED split, the Qwen2.5-VL-7B baseline reaches 7.21\% SR, whereas our SPO model attains 12.02\% SR. The gain, achieved with minimal tuning, suggests that our structured-preference signal transfers beyond ExtendaBench.

\subsection{Visualization}
\begin{figure*}[h]
    \centering
    \includegraphics[width=0.8\linewidth]{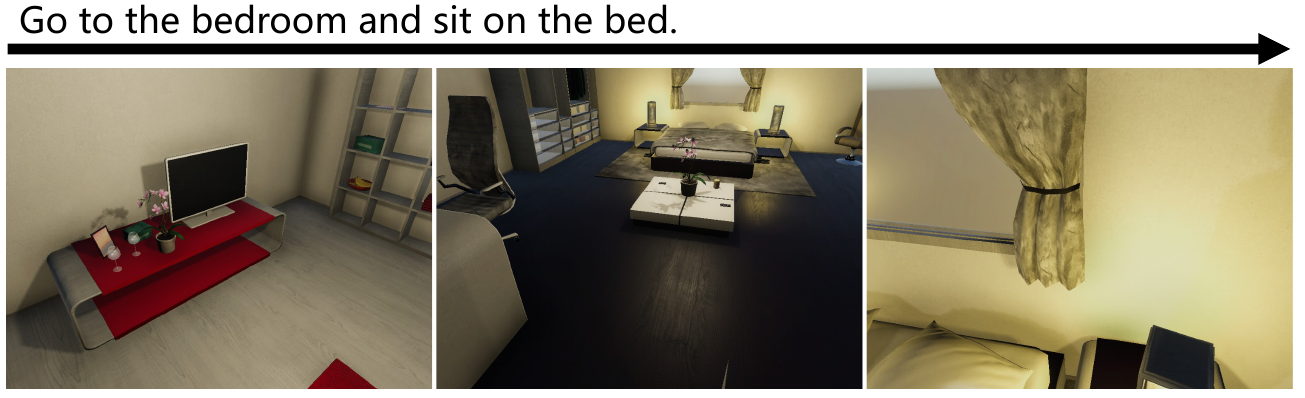}
    \caption{Generated task example in VirtualHome (ultra short).}
    \label{fig:vh_data_ultra_short}
\end{figure*}

\begin{figure*}[h]
    \centering
    \includegraphics[width=\linewidth]{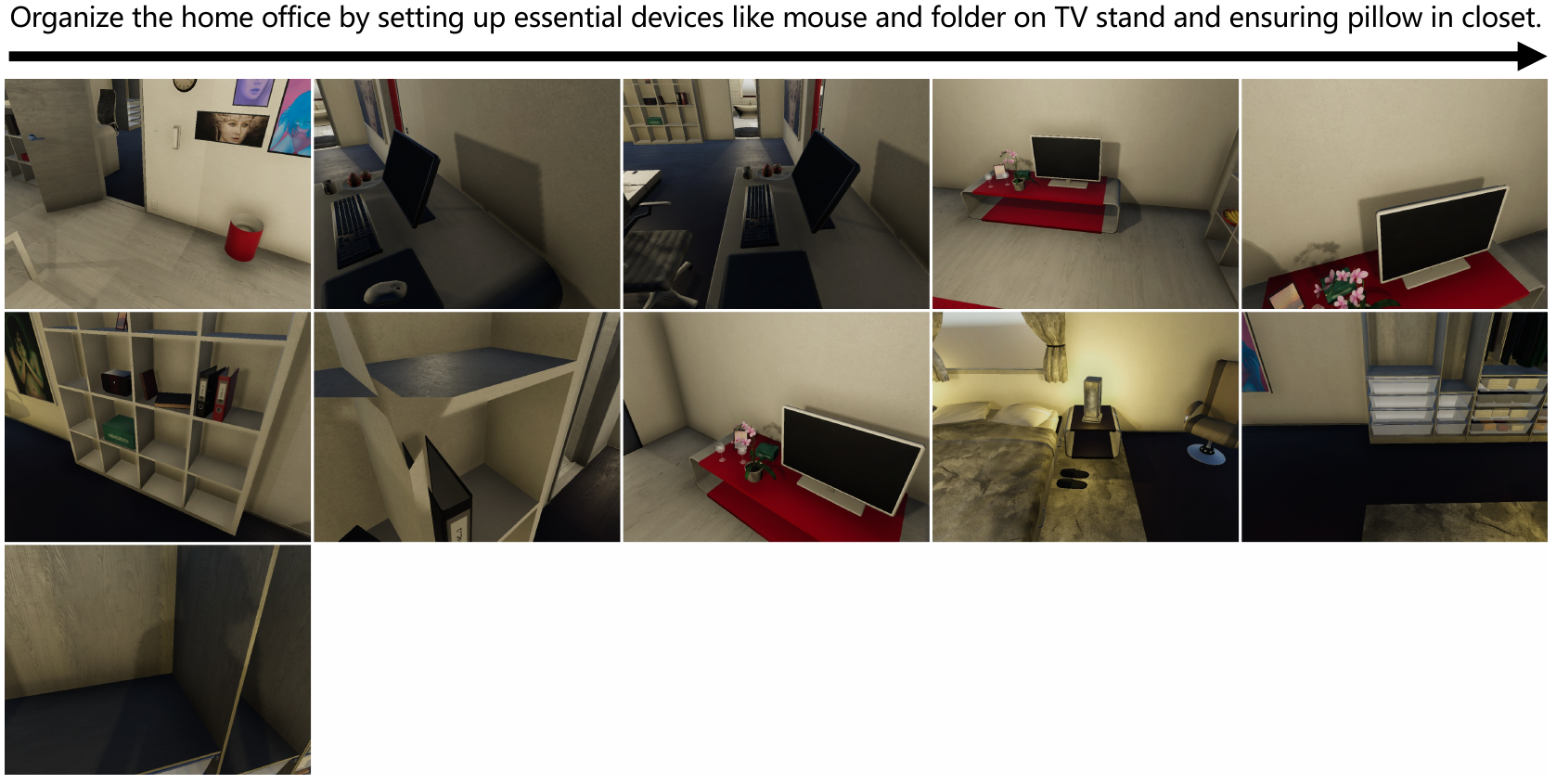}
    \caption{Generated task example in VirtualHome (short).}
    \label{fig:vh_data_short}
\end{figure*}

\begin{figure*}[h]
    \centering
    \includegraphics[width=\linewidth]{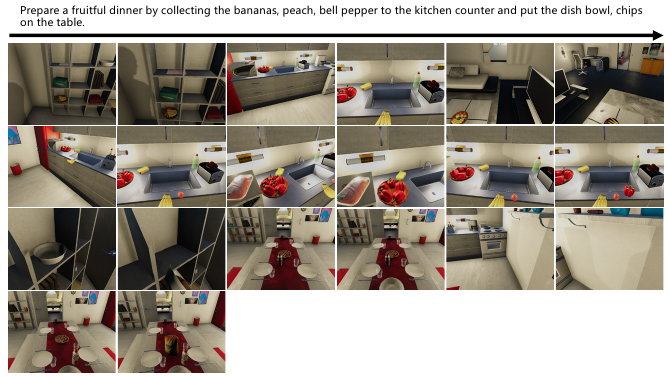}
    \caption{Generated task example in VirtualHome (medium).}
    \label{fig:vh_data_medium}
\end{figure*}

\begin{figure*}[h]
    \centering
    \includegraphics[width=\linewidth]{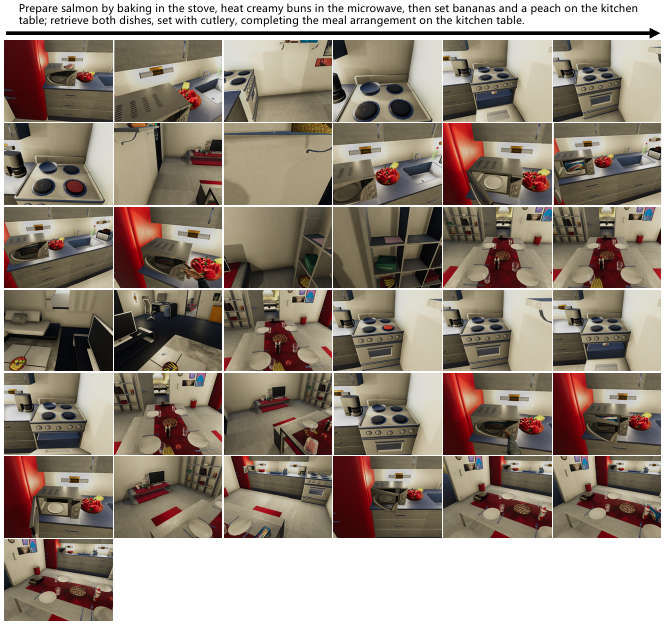}
    \caption{Generated task example in VirtualHome (long).}
    \label{fig:vh_data_long}
\end{figure*}

\begin{figure*}[h]
    \centering
    \includegraphics[width=\textwidth]{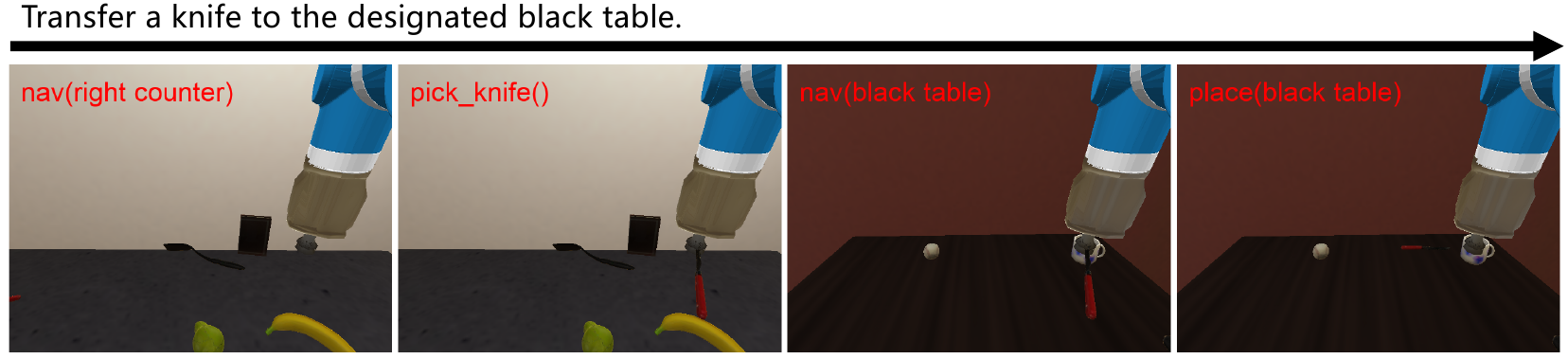}
    \caption{Generated task example in Habitat (ultra short).}
    \label{fig:habitat_data_ultra_short}
\end{figure*}

\begin{figure*}[h]
    \centering
    \includegraphics[width=\textwidth]{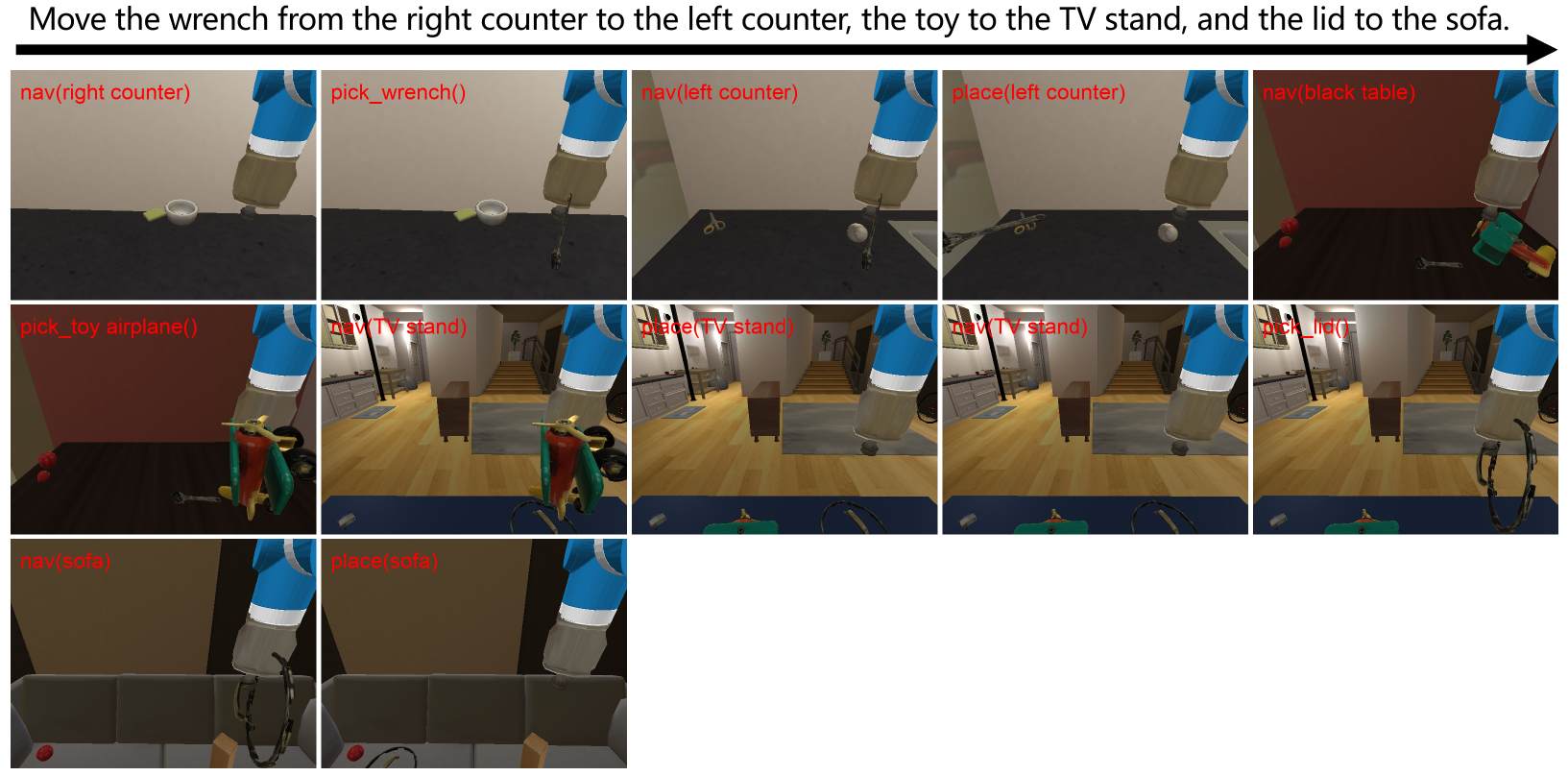}
    \caption{Generated task example in Habitat (short).}
    \label{fig:habitat_data_short}
\end{figure*}

\begin{figure*}[h]
    \centering
    \includegraphics[width=\textwidth]{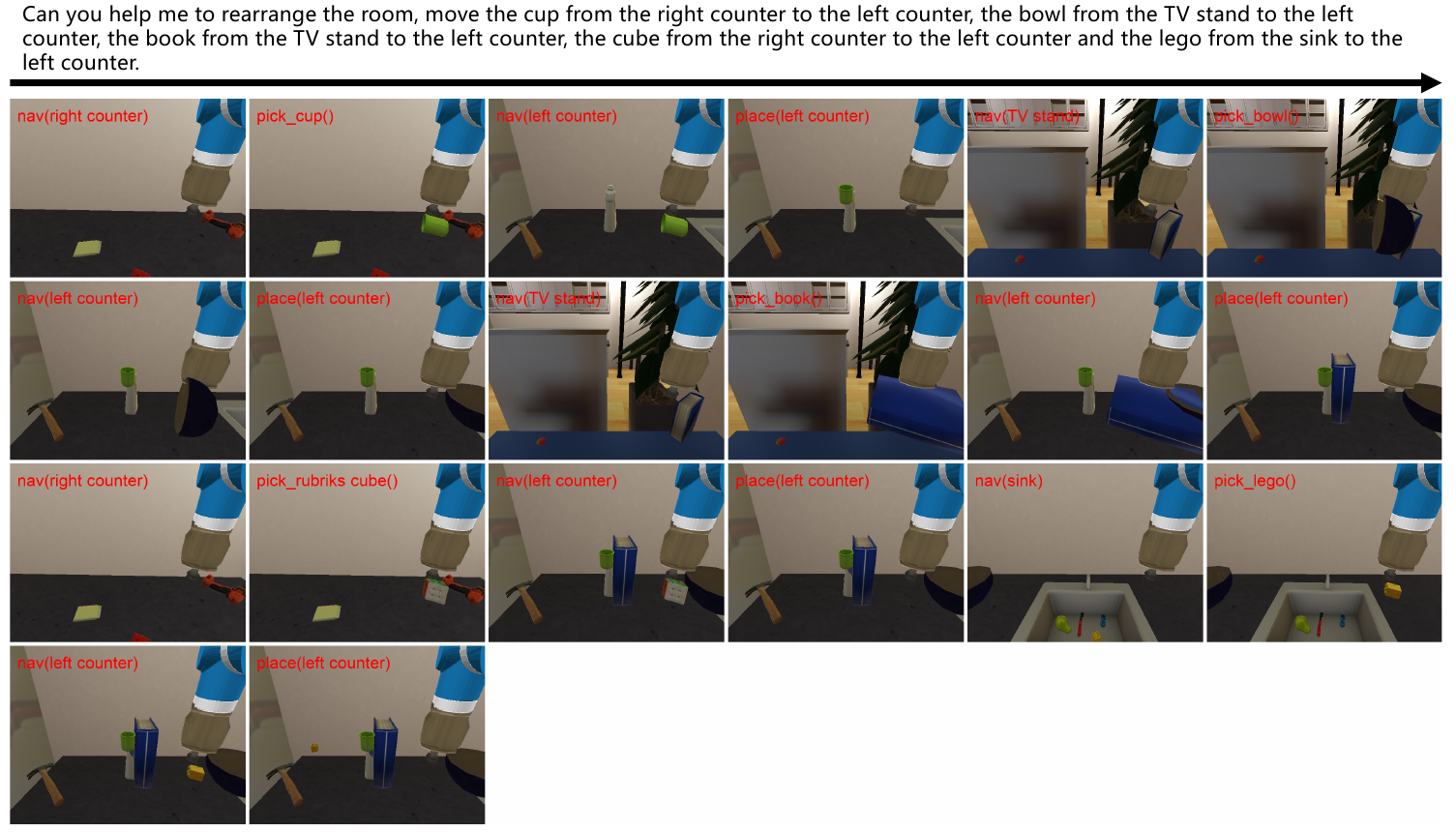}
    \caption{Generated task example in Habitat (medium).}
    \label{fig:habitat_data_medium}
\end{figure*}

\begin{figure*}[h]
    \centering
    \includegraphics[width=\textwidth]{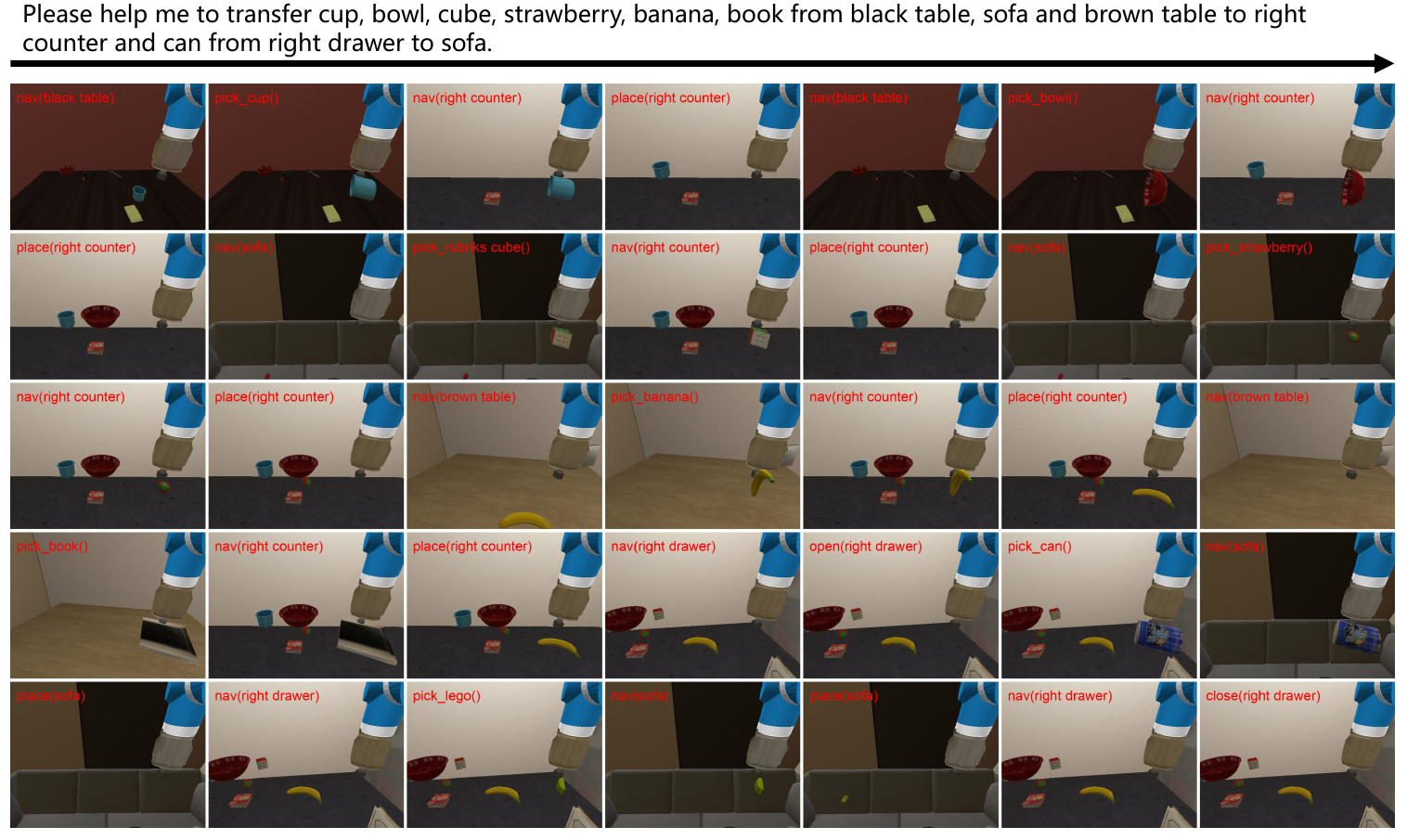}
    \caption{Generated task example in Habitat (long).}
    \label{fig:habitat_data_long}
\end{figure*}

To showcase the diversity and progressive difficulty of tasks in VirtualHome and Habitat 2.0, we present representative visualizations across four difficulty levels: ultra-short, short, medium, and long (Figures~\ref{fig:vh_data_ultra_short}-\ref{fig:vh_data_long}, \ref{fig:habitat_data_ultra_short}-\ref{fig:habitat_data_long}). Tasks are categorized based on action sequence length---a practical proxy for planning complexity.

Figures~\ref{fig:vh_data_ultra_short}-\ref{fig:vh_data_long} illustrate VirtualHome tasks ranging from simple navigation to complex, multi-step meal preparation. Figures~\ref{fig:habitat_data_ultra_short}-\ref{fig:habitat_data_long} show corresponding Habitat tasks that progress from basic object transfers to extensive spatial rearrangements.

\begin{figure*}[h]
    \centering
    \includegraphics[width=\textwidth]{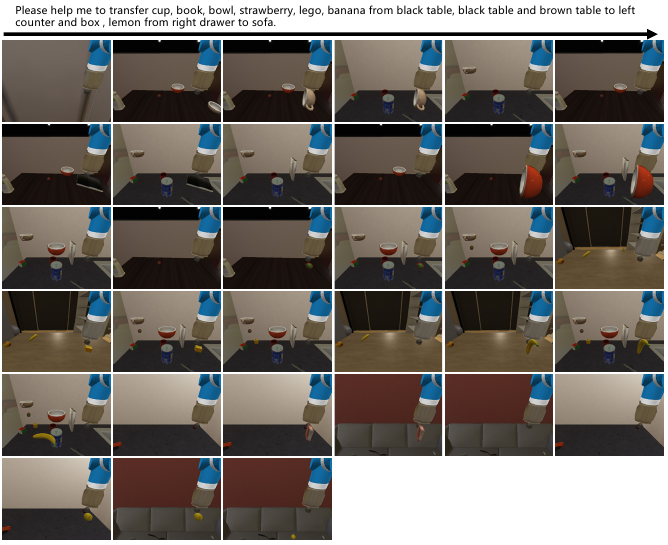}
    \caption{Generated task example in Habitat.}
    \label{fig:habitat_data1}
\end{figure*}

\begin{figure*}[h]
    \centering
    \includegraphics[width=\textwidth]{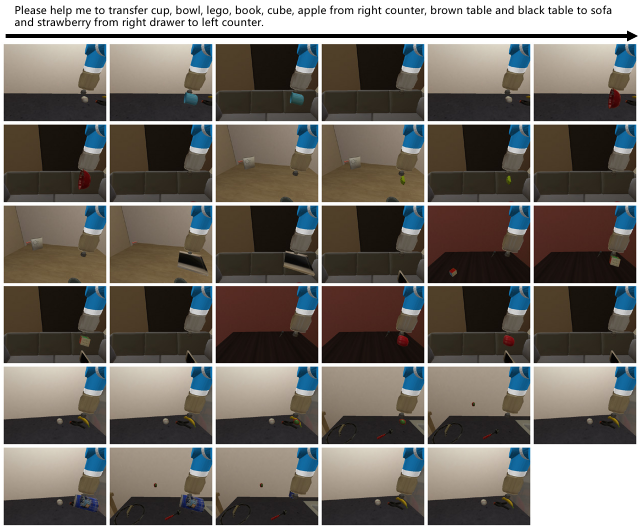}
    \caption{Generated task example in Habitat.}
    \label{fig:habitat_data2}
\end{figure*}

To further highlight the challenges of long-horizon reasoning, we include two additional long task examples in Habitat (Figures~\ref{fig:habitat_data1} and~\ref{fig:habitat_data2}), featuring diverse object types, complex layouts, and longer planning sequences.

These visualizations confirm that our benchmark enables structured, fine-grained evaluation across a spectrum of embodied reasoning difficulties.

\subsection{Case Study}
\begin{figure}[h]
    \centering
    \includegraphics[width=0.9\linewidth]{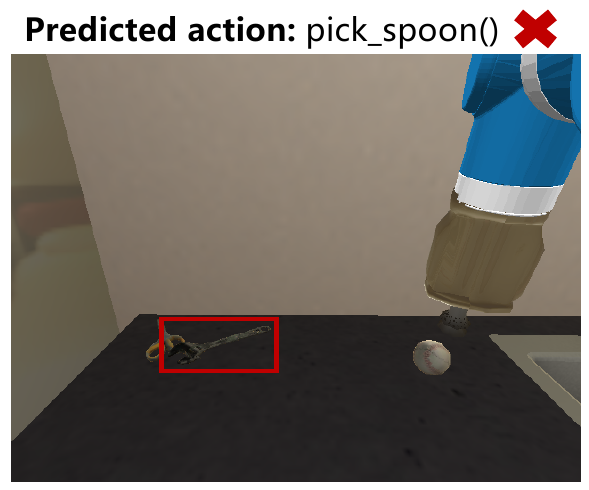}
    \caption{Example of visual grounding failure due to object misidentification.}
    \label{fig:visual_failure}
\end{figure}

Visual grounding failures occasionally occur due to the limitations of the small VLM backbone (Qwen2.5-VL 7B). Common issues include misidentifying small or partially occluded objects (e.g., mistaking a wrench for a spoon as in Figure \ref{fig:visual_failure}) or failing to attend to relevant scene regions. Our structured scoring mechanism---based on textual coherence and image awareness---encourages better alignment between reasoning and visual input, which helps mitigate such errors during training.

\begin{figure}[h]
    \centering
    \includegraphics[width=\linewidth]{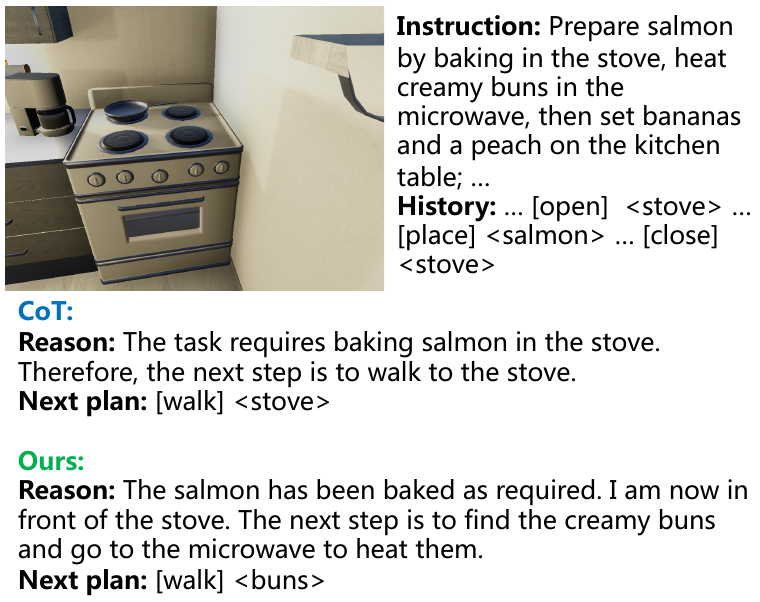}
    \caption{Comparison between baseline CoT and our method on reasoning consistency.}
    \label{fig:reason_vis}
\end{figure}

A common failure in long-horizon tasks is historical inconsistency---where the model fails to consider previously completed steps. As shown in Figure~\ref{fig:reason_vis}, the CoT baseline incorrectly suggests walking to the stove again, ignoring that the salmon has already been baked. In contrast, our method correctly interprets the execution history and proceeds to the next relevant subtask. This demonstrates more coherent and context-aware reasoning.

\section{Prompts}

\subsection{Prompts for Generating Data}

\subsubsection{VirtualHome}

\begin{tcolorbox}[mybox, breakable, title=Task Proposal]
Follow these steps to generate your answer:

1. Think about the task generation:

   - Design a task with more than 30 sequential steps.
   
   - Use only actions from the ``HUMAN ACTION LIST'' and objects from the ``OBJECT LIST.''
   
   - Ensure the task involves at least 12 distinct objects from the ``OBJECT LIST.''

2. Provide a detailed task description:

   - Output a comprehensive description of the task.
   
   - Include all subtasks and the required objects.

3. Decompose the task step by step:

   - Break the task into individual steps.
   
   - After completing each step, analyze and output what needs to be done next.
   
   - Include reasoning for each subsequent step before outputting it.

   Important rules:
   
   - You have only two hands. Each time you grab an object, one hand becomes unavailable until you put the object back.
   
   - Track the number of free hands after each action. Ensure you have at least one free hand before interacting with any object.
   
   - Use only actions from the ``HUMAN ACTION LIST'' and objects from the ``OBJECT LIST.''
   
   - The task must maintain a strong sequential relationship between its decomposed steps, ensuring logical and coherent progression.

\end{tcolorbox}

\begin{tcolorbox}[mybox, breakable, title=Review]
Follow these steps to verify the given task and decomposed steps step by step.\\

- Think about whether the task description is detailed enough to make it clear to a household agent what needs to be done, including every objects in decomposed steps. Give your reasons for this as well as your answer, if the answer is no, give a more detailed description of the task.

- Think and output the reasons why each step is necessary to complete the task.

- Think and output that each step is coherent with a necessary back-and-forth relationship between them.

- Think and output the reasons why the decomposed steps accomplish the task.

- verify the actions in decomposed steps only come from ``HUMAN ACTION LIST.''

- verify the objects in decomposed steps only come from ``OBJECT LIST.'' The inclusion of any additional objects or locations is strictly prohibited.

- Think and output the reasons why each step make common sense.

- verify that each step is compliant with the rule of [walk] object before interacting with it.\\

If the verification passes, return true, otherwise return false and then give your adjustment.
\end{tcolorbox}

\begin{tcolorbox}[mybox, breakable, title=Refinement]
This is the feedback and observation based on your steps that have been executed:

[feedback]

\textless image\textgreater\\

Please perform the following steps based on the feedback:

1. Please think about and output the reason why the steps failed to execute.

2. Based on the reasons why the steps failed, think about and output the reasons why this task is feasible given the rules, and output yes or no.

3. if the task is feasible, output your modifications to the failed step.
\end{tcolorbox}

\subsubsection{Habitat}

\begin{tcolorbox}[mybox, breakable, title=Template Proposal]
You are a robot task generator that can generate robot task templates of different lengths based on given robot actions and examples.\\

The actions you can use include: 

1.nav(obj or receptacle) is used by the robot to navigate to the corresponding object or receptacle 

2.pick(obj) is used by the robot to grab an object 

\ldots\\

Rules:

1. You need to output five parts, including instructions, task planning, replaceable objects and target states. 

2. If the object or receptacle in the instructions and task planning can be replaced, use \` plus pronouns to replace it. 
\end{tcolorbox}

\begin{tcolorbox}[mybox, breakable, title=Instruction Augmentation]
You are a task instruction rewriter, and you can rewrite and expand the robot's task instructions according to the given rewriting rules.\\

Rules:

1. You can use the verbs of the task instructions Use synonyms to replace, for example, change move to reposition.

2. You can replace the objects used in the task instructions, replace the objects with corresponding colors or appearance descriptions, such as changing apple to a red round Fruit.

3. Add some context descriptions, for example, in ``Please put an apple on the table for me,'' change it to ``I want to eat an apple, please put an apple on the table for me'' to make the instruction longer.\\

Now Please help me rewrite the following instructions:
\end{tcolorbox}

\subsection{Prompts for Preference Evaluation}
\label{sec:prompt_preference_evaluation}
\begin{tcolorbox}[mybox, breakable, title=Preference Evaluation]
You are an evaluation system designed to assess how well a reasoning chain (CoT) aligns with the task instruction and how effectively it utilizes the current image observation.\\

Given a task instruction, a reasoning chain (CoT), past execution history, and an RGB image observation, your task is to evaluate:

1. **Task Alignment Score**: How well the reasoning chain follows the task instruction and previous history.

2. **Image Utilization Score**: How well the reasoning chain leverages the current image observation to infer the next step.

3. **Overall Score**: A final score that summarizes the overall quality of the reasoning chain, considering both task alignment and image utilization.\\

**Input Data**:

- **Task Instruction:** \{INSTR\}

- **Chain of Thought (CoT):** \{REASON\}

- **Previous Execution History:** \{HISTORY\}

- **Current Image Observation (RGB):** <image>\\

**Output Format:**

Return the three scores in the following format:

Task Alignment Score: X

Image Utilization Score: Y

Overall Score: Z

Where **X**, **Y**, and **Z** are numbers between **0 and 1**.
\end{tcolorbox}

\section{License}
The dataset is published under CC BY-NCSA 4.0 license, which means everyone can use this dataset for non-commercial research purposes.